\documentclass{article} 

\usepackage{iclr2027_conference,times}

\usepackage[utf8]{inputenc} 
\usepackage[T1]{fontenc}    
\usepackage{hyperref}       
\usepackage{url}            
\usepackage{booktabs}       
\usepackage{graphicx}       
\usepackage{amsmath}        
\usepackage{amsfonts}       
\usepackage{nicefrac}       
\usepackage{microtype}      
\usepackage{xcolor}         
\usepackage{colortbl}       
\usepackage{multirow}       
\usepackage{placeins}
\usepackage{float}
\usepackage{wrapfig}
\usepackage{caption}        
\usepackage[rightcaption,ragged]{sidecap}
\usepackage{pgfplots}
\pgfplotsset{compat=1.18}
\usetikzlibrary{shapes.geometric}
\newcolumntype{I}{!{\vrule width 0.4pt}}

\pgfdeclareplotmark{fivestar}{%
  \node[star,star points=5,star point ratio=2.2,fill=yellow!92!orange,
        draw=black,line width=0.3pt,inner sep=1.25pt,anchor=center]{};%
}
\pgfdeclareplotmark{fivestarsmall}{%
  \node[star,star points=5,star point ratio=2.2,fill=yellow!92!orange,
        draw=black,line width=0.22pt,inner sep=0.7pt,anchor=center]{};%
}
\pgfplotsset{
  ssdpanel/.style={
    width=0.35\textwidth, height=0.21\textwidth,
    xlabel={Latency speedup (\texttimes)},
    tick label style={font=\fontsize{7.5}{9}\selectfont, /pgf/number format/assume math mode=true},
    label style={font=\fontsize{7.5}{9}\selectfont},
    xlabel style={yshift=2pt},
    ylabel style={yshift=-1pt},
    title style={font=\fontsize{7.5}{9}\selectfont\bfseries, yshift=-3pt},
    grid=major, major grid style={draw=gray!20},
    axis line style={draw=gray!60}, tick style={draw=gray!60},
    line width=0.7pt, mark size=1.8pt,
    scaled y ticks=false,
  },
  bAR/.style     ={only marks, mark=square*,   mark size=2.1pt, color=black},
  bSJD/.style    ={only marks, mark=pentagon*, mark size=2.1pt, color=violet!75!black},
  bGSD/.style    ={only marks, mark=diamond*,  mark size=2.4pt, color=teal!85!black},
  bLAN/.style    ={only marks, mark=*,         mark size=1.8pt, color=brown!80!black},
  bZIP/.style    ={only marks, mark=x,         mark size=2.8pt, line width=1.0pt, color=red!80!black},
  ssdcurve/.style={color=blue!70!black, thick, mark=o, mark options={fill=white,draw=blue!70!black}},
  ssdstar/.style    ={only marks, mark=fivestar},
  ssdstarleg/.style ={only marks, mark=fivestarsmall},
}

\usepackage{xcolor}

\definecolor{avgcol}{RGB}{244,235,250}     
\definecolor{ourscol}{RGB}{230,242,255}    
\definecolor{avgcolours}{RGB}{226,228,251} 
\newcommand{\model}{SSD}
\newcommand{\modelfull}{Spatially Speculative Decoding}

\title{\model{}: \modelfull{} Accelerates Autoregressive Image Generation}

\author{%
  Shilong Xiang \quad
  Zirui Zhang \quad
  Lijun Yu\thanks{Equal advising.} \quad
  Chengzhi Mao\footnotemark[1] \\[4pt]
  Rutgers University \\[2pt]
  \texttt{\{shilong.xiang, zirui.zhang, lijun.yu, chengzhi.mao\}@rutgers.edu}\\
  \vspace{2mm}
  \href{https://shilongxiang.github.io/SSD/}{\texttt{https://shilongxiang.github.io/SSD/}}
}

\iclrpreprintcopy 

\begin{document}

\setlength{\tabcolsep}{2pt}
\maketitle
\setlength{\tabcolsep}{6pt}



\begin{abstract}
Autoregressive image models treat images as 1D token sequences, inheriting the next-token factorization of language models. This flattening discards a useful property of images: visual tokens are organized and correlated in two-dimensional space. We introduce \modelfull{} (\model{}), an inference-time decoding framework that exploits this spatial structure for accelerated generation. Rather than speculating only along the flattened sequence, \model{} predicts both the adjacent horizontal token and the token directly below it, allowing multiple spatial directions to advance in parallel. This reduces the number of backbone forward evaluations and alleviates the memory bottleneck of autoregressive decoding. \model{} accelerates image generation by up to 11.03× in wall-clock time while maintaining generation quality on DPG-Bench and GenEval. These results show that spatial structure provides a simple and effective source of parallelism for faster autoregressive image generation.

\end{abstract}

\section{Introduction}
\label{sec:intro}

The appeal of autoregressive visual generation lies in its unification: a single ``next-token'' objective modeling both human text and the visual world~\citep{wu2025janus, chen2025januspro, sun2023emu, xie2024show}, unlocking the potential for multimodal chain-of-thought reasoning and beyond~\citep{zhang2023multimodal}. However, this modality-agnostic architecture introduces a geometric compromise. To fit the linear processing stream of language models, the two-dimensional fabric of images is flattened into a one-dimensional sequence~\citep{team2024chameleon, openai2025gpt4o}. While this linguistic formulation allows vision and text to share a statistical engine, it discards the spatial priors of the physical world~\citep{esser2021taming, ramesh2021dalle, yu2022scaling, sun2024autoregressive, wang2024emu3}. Synthesizing a single image requires generating thousands of discrete visual tokens, making strict next-token prediction inherently slow. The length of these flattened sequences creates a severe inference bottleneck for unified autoregressive models.

A large body of work attempts to accelerate this bottleneck by importing natural language processing techniques. Grafting speculative decoding \citep{leviathan2023fast, chen2023accelerating, li2024eagle, cai2024medusa, li2024eagle2} and Jacobi-based iteration \citep{fu2024break, santilli2023accelerating} onto visual models yields modest 1.8$\times$ to 3.7$\times$ speedups \citep{jang2024lantern, so2025gsd, teng2024sjd, peruzzo2026mulosd, li2026coolsd}. However, these adaptations remain bound to the 1D sequential assumption. Unlike text, visual patches exhibit high localized entropy without rigid grammar; anticipating multiple tokens along a flattened 1D horizon thus yields low draft acceptance rates \citep{jang2024lantern, so2025gsd}. Other methods introduce spatial parallelism, but often rely on restrictive independence assumptions \citep{he2024zipar, wang2024parallelized} or require training new architectures from scratch \citep{he2025nar, zhang2025lpd, li2025arpg}. Recent approaches relax exact token-level equivalence during speculative decoding \citep{he2024zipar,jang2024lantern, li2026coolsd}, while maintaining the generation quality.



\begin{figure}[t]
\centering
\begin{minipage}[c]{0.78\textwidth}
\includegraphics[width=\textwidth,trim=16 0 16 0,clip]{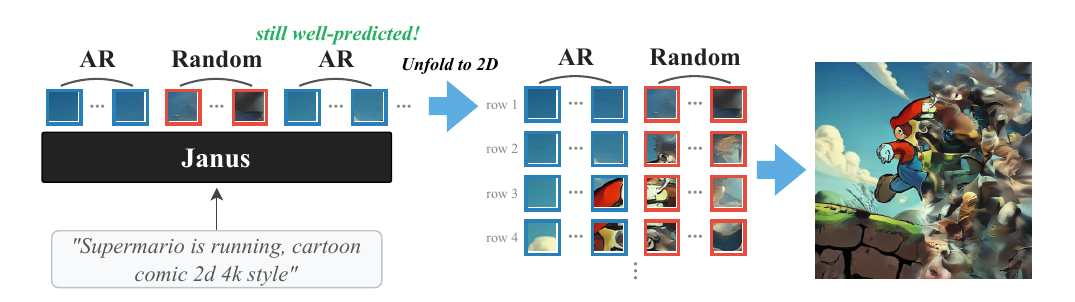}
\end{minipage}%
\hfill
\begin{minipage}[c]{0.20\textwidth}
\centering
\footnotesize
\setlength{\tabcolsep}{3pt}
\textbf{\scriptsize Accept Rate (\%)}\\[3pt]
\begin{tabular}{@{}c cc@{}}
\toprule
Off. & Horiz. & Vert. \\
\midrule
$+1$ & 43.27 & 30.32 \\
$+2$ & 29.69 & 27.03 \\
$+3$ & 25.47 & 25.71 \\
\bottomrule
\end{tabular}
\end{minipage}
\caption{\textbf{The Two-Dimensional Nature of Predictive Dependency.} To demonstrate that spatial correlations are inherently 2D, we corrupt the sequential context during Janus-Pro-7B generation by replacing the second half of each row with random tokens (red outlines). Despite this severe disruption to the 1D sequence, visual coherence is preserved wherever the token directly \emph{above} was accurately generated (blue outlines). This confirms that vertical prediction relies fundamentally on spatial adjacency rather than position in the flattened raster-scan order. \textbf{(Right)}~Acceptance rates of horizontal vs.\ vertical drafting heads at matching spatial offsets, confirming that predictability is governed by 2D spatial locality.}
\label{fig:vertical_probe}
\end{figure}

\begin{figure}[t]
\centering
\includegraphics[width=\textwidth,trim=20 10 15 6,clip]{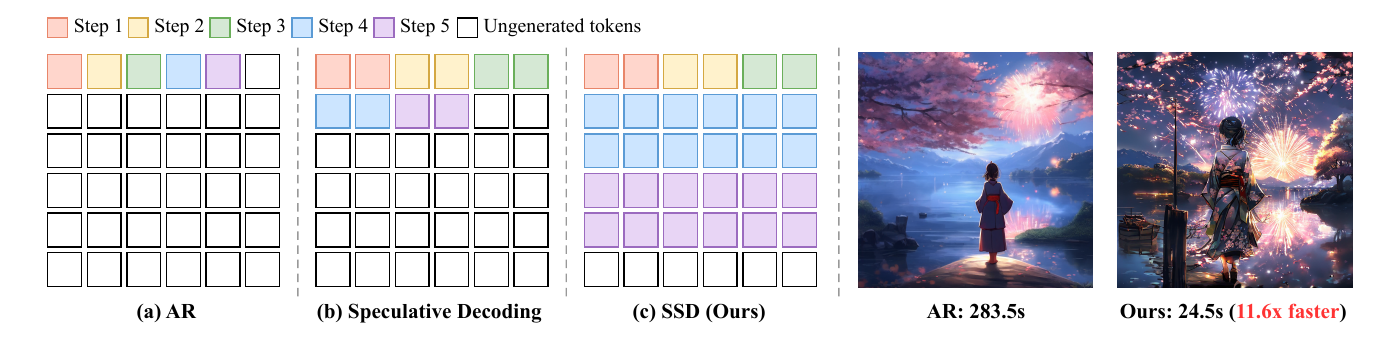}
\caption{\textbf{Accelerating Autoregressive Vision via 2D Spatial Anticipation.} (a) Standard AR flattens the visual world into a 1D sequence and predicts one token per backbone forward evaluation. (b) Speculative Decoding accelerates generation locally but remains constrained by this linear raster-scan geometry. (c) Our \model{} aligns the predictive objective with the intrinsic geometry of images and drafts entire spatial blocks in parallel. This reduces the number of backbone forward evaluations and alleviates the memory bottleneck of autoregressive decoding. Right: Applied to Emu3 (8B), this geometric shift yields a 11.03$\times$ speedup while preserving high-resolution visual fidelity.
}
\label{fig:teaser}
\end{figure}

In this paper, we introduce \modelfull{} (\model{}), a framework that aligns the mechanics of autoregressive generation with the intrinsic structure of vision domain. Our key insight is that spatial correlations are inherently two-dimensional: within a raster-scan sequence, the token directly below a position shares a predictive dependency comparable to the immediate next token (Figure~\ref{fig:vertical_probe}). Rather than guessing a fragile 1D future, our approach simultaneously anticipates visual space across horizontal and vertical axes (Figure~\ref{fig:teaser}). By respecting this 2D locality, the model drafts entire rows at once. This geometric shift reduces the number of forward evaluations (NFE) complexity, which is the key bottleneck for inference speed.


We further reduce verification cost by following the non-exact acceptance schemes used in visual speculative decoding~\cite{he2024zipar,jang2024lantern, li2026coolsd}. \model{} drafts many spatially correlated tokens at once, so truncating the draft at the first mismatch can discard many useful predictions. We instead use the verification pass to correct the draft: mismatched tokens are replaced with predictions from the backbone, and the resulting block can be verified again. Repeating this correction gives a simple knob between speed and adherence to the original model distribution. Empirically, two verification passes already produce images visually indistinguishable from standard autoregressive decoding (Figure~\ref{fig:verify_samples}).

A key advantage of our framework is that it factors out inference acceleration from backbone pretraining. Because \model{} requires no modifications to the pretrained backbone, it serves as a plug-and-play module that accelerates visual generation when needed and leaves the original model untouched when removed. Moreover, the modularity of our approach allows it to generalize to any unified autoregressive model that produces discrete visual tokens. Because these drafting heads are lightweight, the FLOPs overhead is minimal (Figure~\ref{fig:draft_cost}), enabling massive computational savings by successfully overcoming the memory wall.

\begin{figure}[t!]
\centering
\includegraphics[width=\textwidth]{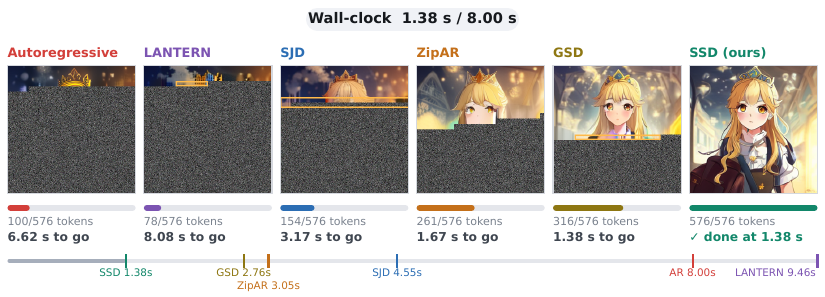}
\caption{\textbf{Generation Progress at a Matched Wall-Clock Budget.} The snapshots show Janus-Pro-7B generation for the same prompt at the instant \model{} finishes (1.38\,s): \model{} has completed all 576 image tokens, whereas autoregressive decoding and existing state-of-the-art acceleration methods remain largely incomplete. The bottom axis marks the total latency at which each method finishes. The snapshots and completion markers provide an intuitive illustration of the substantial speedup achieved by \model{} over all baselines.}
\label{fig:nfe_snapshot}
\end{figure}

Figure~\ref{fig:nfe_snapshot} illustrates this speedup with a single generation example on Janus-Pro. At the moment \model{} completes all 576 image tokens, autoregressive decoding and existing acceleration methods have produced only partial images. Across the full evaluation, \model{} accelerates wall-clock generation by 7.35$\times$ on Lumina-mGPT~\citep{liu2024lumina}, 5.70$\times$ on Janus-Pro~\citep{chen2025januspro}, and 11.03$\times$ on Emu3~\citep{wang2024emu3}. Our results show that 1D sequential decoding is a costly artifact of the language-model formulation rather than a necessity for image generation. By exploiting the geometry of images, \model{} removes much of this unnecessary sequential computation.


\section{Related Work}
\label{sec:related_work}

\noindent\textbf{Autoregressive Image Generation.}
Discrete visual tokenization approaches, such as VQ-VAE~\citep{van2017neural}, VQGAN~\citep{esser2021taming}, and MAGVIT-v2~\citep{yu2024magvit2}, allow images to be represented as discrete latent features. This discrete representation enables standard transformer architectures to synthesize images via autoregressive next-token prediction~\citep{ramesh2021dalle, yu2022scaling, sun2024autoregressive}. Furthermore, this shared discrete formulation has catalyzed the development of unified vision-language models~\citep{wu2025janus, chen2025januspro, sun2023emu, xie2024show, team2024chameleon, wang2024emu3} capable of performing both understanding and generation within a single backbone. 


\noindent\textbf{Multi-Token Prediction and Speculative Decoding.}
Multi-token prediction (MTP)~\citep{gloeckle2024better} has emerged as an effective strategy to bypass this memory wall. By utilizing auxiliary heads to anticipate multiple future tokens during a single forward pass, MTP significantly reduces the number of required backbone forward passes. To ensure the output quality of these MTP drafts matches the target distribution, speculative decoding~\citep{leviathan2023fast, chen2023accelerating} verifies candidate tokens in a single parallel pass~\citep{cai2024medusa, li2024eagle, li2024eagle2, li2025eagle3, miao2024specinfer}. Because this verification is computed in parallel across all draft tokens, it successfully overcomes the memory wall by requiring only a single forward pass for verification. Similarly, Jacobi-based methods~\citep{fu2024break, santilli2023accelerating, kou2024cllms} build upon this parallel verification principle by framing decoding as fixed-point convergence achieved via iterative forward passes, thereby eliminating the need for a separate draft model. While these techniques yield speedups in text generation, their application to visual domains is limited by the one-dimensional sequential setting, as they only anticipate subsequent tokens along a flattened 1D raster-scan order.

\noindent\textbf{Accelerating Image Autoregressive Generation.}
Consequently, directly applying these 1D-centric acceleration techniques to visual autoregressive models~\citep{jang2024lantern, so2025gsd, teng2024sjd, peruzzo2026mulosd, li2026coolsd} yields marginal improvements, typically plateauing at $1.8\times$ to $3.7\times$ speedups. Recent methods explore spatial parallelism under two distinct settings. Existing plug-in acceleration methods typically keep the pretrained backbone frozen and introduce only lightweight auxiliary prediction heads, but achieve limited speedups~\citep{jang2024lantern, so2025gsd, teng2024sjd, peruzzo2026mulosd, li2026coolsd}. Other methods~\citep{he2025nar, wang2024parallelized, zhang2025lpd, li2025arpg, zhou2026flashar} pursue stronger parallelism by modifying and retraining the backbone. While effective, backbone retraining is costly for large foundation models and may alter their pretrained capabilities. PJD~\citep{liao2026parallel} instead extends Jacobi decoding to two-dimensional row-wise refinement without learning spatial drafts. In contrast, our SSD preserves the pretrained backbone and learns only lightweight spatial drafting heads, while exploiting 2D spatial structure to achieve substantially larger acceleration.

\section{Method: Spatially Speculative Decoding}

\subsection{Autoregressive Visual Generation in Unified Vision Language Models}
\label{sec:formulation}

Modern autoregressive image models~\citep{ramesh2021dalle, yu2022scaling, sun2024autoregressive} typically consist of two components: a vector quantizer (VQ)~\citep{van2017neural, esser2021taming} that discretizes images into tokens, and an autoregressive transformer~\citep{vaswani2017attention} that models the resulting sequence. The VQ encoder $\mathcal{E}$ maps a continuous image $\mathbf{X} \in \mathbb{R}^{H \times W \times 3}$ to a grid of latent features, each quantized to its nearest entry in a learned codebook $\mathcal{V}$, yielding a discrete token grid $\mathbf{T} \in \mathcal{V}^{n \times n}$. A decoder $\mathcal{D}$ reconstructs the image from $\mathbf{T}$.

Given this discrete representation, image generation reduces to sequential token prediction. The grid $\mathbf{T}$ is flattened into a one-dimensional sequence $(t_1, t_2, \ldots, t_{n^2})$ via raster-scan ordering. In unified vision-language models~\citep{wu2025janus, chen2025januspro, team2024chameleon}, visual and text tokens share the same vocabulary and transformer backbone, enabling autoregressive generation:
\begin{equation}
    t_i \sim p_\theta(t_i \mid \mathbf{c}, t_1, \ldots, t_{i-1}),
\end{equation}
where $\mathbf{c}$ is the text prompt prefix. Generating an $n \times n$ image needs $n^2$ sequential forward passes.

\subsection{Accelerating Inference via 1D Speculative Decoding}
\label{sec:2dmtp}

\begin{wrapfigure}{r}{0.38\textwidth}
\vspace{-19pt}
\centering
\includegraphics[width=0.38\textwidth]{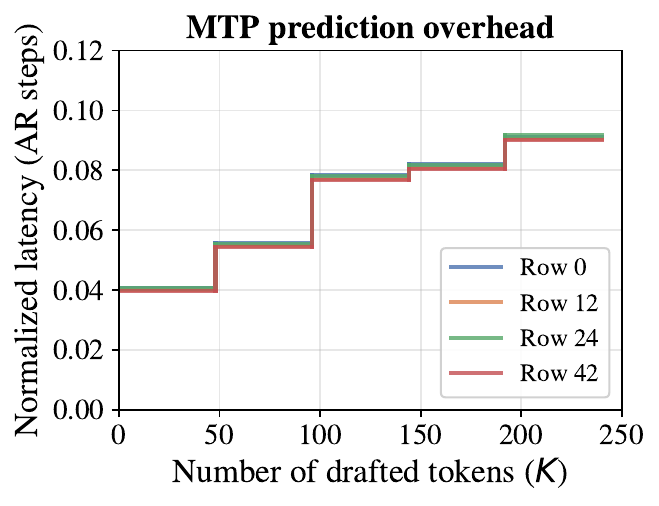}
\vspace{-6pt}
\caption{MTP drafting cost of \model{}, normalized by one AR step on Lumina-mGPT-7B ($48{\times}48$ grid). Even at 240 drafted tokens, overhead stays below $0.1$ AR steps.}
\label{fig:draft_cost}
\vspace{-14pt}
\end{wrapfigure}

Autoregressive (AR) image generation loads billions of parameters from memory at every step yet produces only a single token per forward pass, creating a memory-bandwidth bottleneck~\citep{pope2023efficiently} known as the memory wall. Speculative decoding accelerates generation by first drafting candidate tokens and verifying them in a single forward pass, mitigating the memory wall by significantly reducing the number of times the large language model (LLM) parameters must be reloaded to predict sequential tokens.

Multi-token prediction (MTP)~\citep{gloeckle2024better, deepseekai2024v3, lmtp2025} is a widely used drafting method that adds $K$ lightweight prediction heads to the backbone. These heads, which are vastly smaller than the backbone, predict the next $K$ tokens simultaneously using the original head features with marginal additional computational cost. The drafted tokens are then verified simultaneously by the backbone, requiring one parameter load instead of $K$ sequential loads. As shown in Figure~\ref{fig:draft_cost}, MTP drafting heads are exceptionally lightweight: drafting up to 240 tokens costs less than 0.1 AR steps.

\subsection{Spatially Speculative Decoding}

Standard MTP drafting is inherently one-dimensional, sequentially predicting the immediate next tokens. While suitable for naturally 1D text sequences, 1D drafting along a raster-scan sequence is suboptimal for images because it ignores their two-dimensional structure. This limitation becomes severe for distant spatial neighbors. For instance, predicting the token directly below a given position in a row of width 48 requires a $+48$ offset head in the flattened sequence. Training such long horizon heads for every offset from $+1$ to $+48$ under standard MTP is inefficient, as it requires a separate head for each offset, and accuracy typically degrades with sequence distance.

\begin{wrapfigure}{r}{0.38\textwidth}
\vspace{-12pt}
\centering
\includegraphics[width=0.38\textwidth]{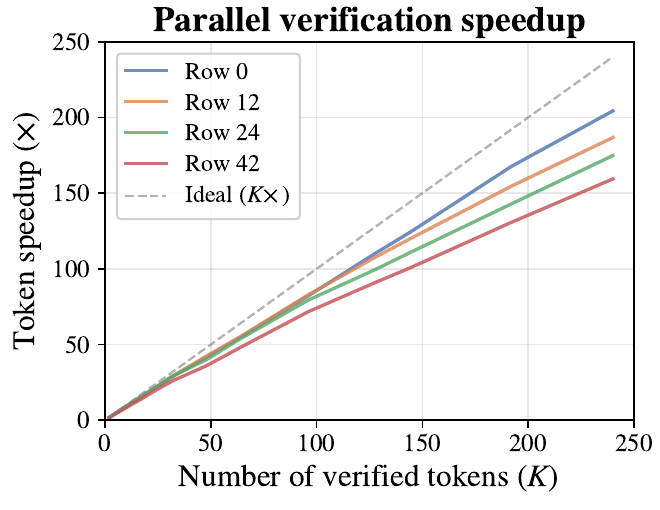}
\vspace{-8pt}
\caption{Token acceleration through parallel verification on Lumina-mGPT-7B. Wall-clock speedup over sequential AR scales near-linearly with the number of simultaneously verified tokens $K$, reaching $205\times$ at $K{=}240$.}
\label{fig:mtp_analysis}
\vspace{-10pt}
\end{wrapfigure}

However, we find that this degradation is an artifact of the raster scan order, not the underlying visual geometry. A token directly below a position is spatially adjacent despite its distance in the flattened sequence. Empirically, vertical heads at matching spatial offsets are just as accurate as their horizontal counterparts: a $+48$ vertical prediction matches a $+2$ horizontal prediction, and a $+96$ vertical prediction matches a $+3$ horizontal prediction (Figure~\ref{fig:vertical_probe}). This observation suggests that MTP should be structured according to image geometry rather than sequence distance. Instead of training heads for arbitrary raster offsets, we train a small, localized set, such as $k_h=3$ horizontal heads for offsets $(+1,+2,+3)$ and $k_v=3$ vertical heads for offsets $(+48,+96,+144)$, bypassing the dense intermediate heads required by standard 1D-MTP.

Vertical prediction also offers a crucial advantage absent in horizontal prediction: column-wise independence. Along a raster-scan row, tokens must be generated sequentially because each token conditions on previous tokens in that same row. In contrast, vertical prediction can draft one token per column directly from the completed row above. Because these predictions do not depend on each other within the target row, they can be produced simultaneously, allowing an entire row of $n$ tokens to be drafted in a single parallel step.

Our method, \emph{\modelfull{}} (\model{}), factorizes 2D anticipation into two orthogonal 1D prediction streams. We first use $k_h$ horizontal heads to draft along the raster-scan direction until the current row is completed. We then apply $k_v$ vertical heads to draft up to $k_v$ subsequent rows in parallel (Figure~\ref{fig:teaser}). The drafted rows are subsequently checked and corrected via \emph{verification as auto correction}. By replacing token-by-token generation within each row with parallel row drafting, \model{} reduces the number of backbone forward evaluations and alleviates the memory bottleneck of autoregressive decoding.

\label{sec:vert_pred}
\paragraph{Draft Prediction in Latent Space} Anticipating in discrete token space demands the exact codebook entry out of tens of thousands of candidates. Visual codebooks induce flat distributions over this vast label set, and exact-match acceptance rates fall below $5\%$~\citep{so2025gsd}. We empirically find that predicting in the continuous latent space---specifically, targeting the last transformer layer's hidden state before the final RMSNorm~\citep{zhang2019root} (see Appendix~\ref{app:target_space} for ablations)---produces better MTP accuracy. A lightweight predictor $f_\phi$ takes as input the hidden state $\mathbf{h}_{y,x}$ concatenated with the token embedding $\mathbf{e}_{y,x}$ (to resolve sampling non-determinism~\citep{li2024eagle}) and predicts the hidden state at offset $\delta$:
\begin{equation}
    \hat{\mathbf{h}}_{y+\delta,x} = f_\phi\bigl([\mathbf{h}_{y,x};\; \mathbf{e}_{y,x}]\bigr), \quad f_\phi(\mathbf{z}) = \mathbf{W}_0\mathbf{z} + \mathrm{SwiGLU}\bigl(\mathrm{RMSNorm}(\mathbf{W}_0\mathbf{z})\bigr),
\end{equation}
where $\mathrm{SwiGLU}(\tilde{\mathbf{z}}) = \mathbf{W}_2\bigl(\sigma(\mathbf{W}_1\tilde{\mathbf{z}}) \odot \mathbf{W}_3\tilde{\mathbf{z}}\bigr)$~\citep{shazeer2020glu} with $\sigma$ denoting SiLU~\citep{ramachandran2017searching}, $\mathbf{W}_0 \!\in\! \mathbb{R}^{d \times 2d}$, $\mathbf{W}_1, \mathbf{W}_3 \!\in\! \mathbb{R}^{m \times d}$, $\mathbf{W}_2 \!\in\! \mathbb{R}^{d \times m}$ are bias-free projections. Each offset $\delta$ and spatial direction has its own predictor. Draft tokens are decoded through the base model's existing output layers: $\hat{t}_{y+\delta,x} = \arg\max_v\; \texttt{GenHead}(\texttt{RMSNorm}(\hat{\mathbf{h}}_{y+\delta,x}))_v$. Training minimizes a smooth L1 loss~\citep{girshick2015fast} against ground-truth hidden states from the frozen backbone:
\begin{equation}
\mathcal{L}_\delta = \frac{1}{|\mathcal{P}_\delta|} \sum_{(y,x) \in \mathcal{P}_\delta} \text{SmoothL1}\bigl(\hat{\mathbf{h}}_{y+\delta,x},\; \mathbf{h}_{y+\delta,x}\bigr),
\end{equation}
where $\mathcal{P}_\delta$ is the set of valid source positions. The pretrained model remains entirely unmodified.

\label{sec:verification}

\paragraph{Verification as Auto Correction.}
Once future tokens are drafted, we pass the entire drafted block through the frozen backbone in parallel for verification and correction. Because spatial drafting predicts one or more complete rows at a time, strict speculative verification that rolls back all positions after the first mismatch would largely eliminate the benefit of row-parallel drafting. We instead adopt a relaxed correction rule that repairs mismatched positions in parallel over a fixed number of verification rounds. This type of non-exact distribution preserving acceleration algorithm has been studied in prior work~\cite{he2024zipar,jang2024lantern, li2026coolsd}.

 For each drafted position $k$, the draft distribution $q_k$ produces a candidate token $\hat{t}_k=\arg\max_v q_k(v)$. During the subsequent verification-and-correction forward pass, the LLM backbone produces an updated distribution $p_k$. We consider three strategies for selecting the final token at each drafted position: \textbf{(A) Greedy selection}, which directly takes $\arg\max_v p_k(v)$; \textbf{(B) Sampling}, which samples a token from $p_k$; and \textbf{(C) Acceptance--rejection sampling}, our default strategy. Specifically, we accept the drafted token $\hat{t}_k$ with probability $\alpha(\hat{t}_k)=\min\left(p_k(\hat{t}_k)/q_k(\hat{t}_k),1\right)$. If the draft is rejected, we instead sample from the residual distribution $\tilde{p}_k(v)=[p_k(v)-q_k(v)]_+/\sum_{v'}[p_k(v')-q_k(v')]_+$, where $[x]_+=\max(0,x)$. This correction is applied independently to all drafted positions in parallel. We compare these three strategies in Table~\ref{tab:abl_accept}, where acceptance-rejection sampling achieves the best performance.


We repeat this verify-and-correct procedure for a fixed number of rounds $r$. After each round, rejected positions are replaced by samples from the residual distribution above, and the backbone is re-evaluated on the updated block. For a rejected position, the previous-round backbone distribution becomes its reference distribution in the next round, while accepted positions retain their current reference distributions. Increasing $r$ progressively refines the drafted block toward the autoregressive model at the cost of additional latency, making $r$ an explicit knob for controlling the speed--quality trade-off. After the final round, one additional forward pass commits the corrected block and updates the KV cache for subsequent generation.

\begin{table}[t!]
\centering
\vspace{-5mm}
\caption{Text-to-image generation results on \textbf{DPG-Bench}. For each model, we compare our \model{} with original model's autoregressive inference, and speculative decoding baselines: 1D-MTP, SJD~\citep{teng2024sjd}, GSD~\citep{so2025gsd}, LANTERN~\citep{jang2024lantern}, and ZipAR~\citep{he2024zipar}.  \colorbox{ourscol}{Blue} rows indicate our method. Our approach establishes a stronger speed--quality Pareto frontier, matching or exceeding prior acceleration baselines in generation quality at substantially higher speed.}
\label{tab:dpgbench}
\resizebox{\textwidth}{!}{%
\small
\setlength{\aboverulesep}{0pt}\setlength{\belowrulesep}{0pt}\renewcommand{\arraystretch}{1.18}
\begin{tabular}{lIcIccIcccccI>{\columncolor{avgcol}}c}
\toprule
\multirow{2}{*}{\textbf{Model}} & \multirow{2}{*}{\textbf{Latency ($\downarrow$)}} & \multicolumn{2}{cI}{\textbf{Acceleration ($\uparrow$)}} & \multirow{2}{*}{\textbf{Global}} & \multirow{2}{*}{\textbf{Entity}} & \multirow{2}{*}{\textbf{Attribute}} & \multirow{2}{*}{\textbf{Relation}} & \multirow{2}{*}{\textbf{Other}} & \multicolumn{1}{>{\columncolor{avgcol}}c}{} \\
\cmidrule(lr){3-4}
& & \textbf{Step} & \multicolumn{1}{cI}{\textbf{Latency}} & & & & & & \multirow{-2}{*}{\textbf{Average ($\uparrow$)}} \\
\midrule
\textbf{Lumina-mGPT (7B)}     & 96.23s & $1.00\times$ & $1.00\times$ & 83.28 & 82.79 & 80.89 & 88.11 & 60.98 & 76.30 \\
+ 1D-MTP                       & 49.91s & $2.23\times$   & $1.93\times$   & 76.78 & 73.11 & 74.61 & 83.67 & 56.73 & 63.20 \\
+ SJD                          & 61.08s & $2.23\times$   & $1.58\times$   & 82.30 & 81.88 & 79.06 & 86.96 & 58.78 & 74.03 \\
+ GSD                          & 38.23s & $2.89\times$   & $2.52\times$   & 83.89 & 81.18 & 79.38 & 87.54 & 59.60 & 73.37 \\
+ LANTERN                      & 59.00s & $2.00\times$   & $1.63\times$   & 82.98 & 82.31 & 79.40 & 88.63 & 57.60 & 74.45 \\
+ ZipAR                         & 36.45s & $2.94\times$   & $2.64\times$   & 83.80 & 82.70 & 80.81 & 88.05 & 61.60 & 74.62 \\
\rowcolor{ourscol} + SSD (Ours)                   & \textbf{13.09s} & $\mathbf{8.40\times}$   & $\mathbf{7.35\times}$   & 81.73 & 82.60 & 79.77 & 88.70 & 58.54 & \cellcolor{avgcolours}74.57 \\
\midrule
\textbf{Janus-Pro (7B)}       & 7.87s & $1.00\times$ & $1.00\times$ & 81.06 & 90.32 & 87.47 & 93.41 & 79.27 & 84.23 \\
+ 1D-MTP                       & 3.94s & $2.48\times$   & $2.00\times$   & 79.69 & 87.80 & 86.00 & 92.96 & 72.36 & 80.98 \\
+ SJD                          & 5.25s & $1.81\times$   & $1.50\times$   & 83.33 & 89.44 & 87.36 & 93.83 & 79.59 & 84.01 \\
+ GSD                          & 3.31s & $3.07\times$   & $2.38\times$   & 81.46 & 89.43 & 87.55 & 93.38 & 78.80 & 82.97 \\
+ LANTERN                      & 9.32s & $1.17\times$   & $0.84\times$   & 82.07 & 89.73 & 87.47 & 93.31 & 82.40 & 84.07 \\
+ ZipAR                        & 3.04s & $2.76\times$   & $2.59\times$   & 79.94 & 89.60 & 87.65 & 93.58 & 80.80 & 83.78 \\
\rowcolor{ourscol} + SSD (Ours)           & \textbf{1.38s} & $\mathbf{7.20\times}$   & $\mathbf{5.70\times}$   & 82.61 & 88.90 & 87.33 & 93.46 & 77.64 & \cellcolor{avgcolours}83.40 \\
\midrule
\textbf{Emu3 (8B)}            & 281.87s & $1.00\times$ & $1.00\times$ & 86.02 & 85.69 & 85.26 & 91.89 & 64.23 & 78.69 \\
+ 1D-MTP                       & 144.26s & $2.39\times$   & $1.95\times$   & 72.05 & 67.42 & 62.87 & 85.27 & 46.94 & 53.11 \\
+ SJD                          & 260.41s & $2.24\times$   & $1.08\times$   & 84.83 & 87.00 & 85.28 & 91.55 & 66.94 & 78.38 \\
+ GSD                          & 97.07s & $3.46\times$   & $2.90\times$   & 82.97 & 85.73 & 82.58 & 91.80 & 65.85 & 77.57 \\
+ LANTERN                      & 182.50s & $2.16\times$   & $1.54\times$   & 82.66 & 84.00 & 81.47 & 89.33 & 71.95 & 76.12 \\
+ ZipAR                         & 79.71s & $5.41\times$   & $3.54\times$   & 83.75 & 81.85 & 82.33 & 89.73 & 67.35 & 73.36 \\
\rowcolor{ourscol} + SSD (Ours)           & \textbf{25.55s} & $\mathbf{14.34\times}$   & $\mathbf{11.03\times}$   & 81.48 & 84.31 & 81.59 & 90.35 & 58.30 & \cellcolor{avgcolours}76.03 \\
\bottomrule
\end{tabular}%
}
\vspace{-2mm}
\end{table}

\begin{table}[t!]
\centering
\caption{Text-to-image generation results on \textbf{GenEval}. For each model, we compare our \model{} with the autoregressive (AR) baseline. \textbf{Ours} denotes a configuration that achieves a competitive trade-off between generation quality and inference acceleration. \colorbox{ourscol}{Blue} rows indicate our method. Latencies are measured on GenEval prompts for Janus-Pro and Emu3; for Lumina-mGPT we report the DPG-Bench measurement of the same configuration. Our approach achieves up to $11.24\times$ latency speedup while retaining competitive generation quality.}
\label{tab:geneval}
\resizebox{\textwidth}{!}{%
\small
\setlength{\aboverulesep}{0pt}\setlength{\belowrulesep}{0pt}\renewcommand{\arraystretch}{1.18}
\begin{tabular}{lIcIccIccccccI>{\columncolor{avgcol}}c}
\toprule
\multirow{2}{*}{\textbf{Model}} & \multirow{2}{*}{\textbf{Latency ($\downarrow$)}} & \multicolumn{2}{cI}{\textbf{Acceleration ($\uparrow$)}} & \multirow{2}{*}{\textbf{Single}} & \multirow{2}{*}{\textbf{Two}} & \multirow{2}{*}{\textbf{Count.}} & \multirow{2}{*}{\textbf{Colors}} & \multirow{2}{*}{\textbf{Pos.}} & \multirow{2}{*}{\textbf{Col. Attr.}} & \multicolumn{1}{>{\columncolor{avgcol}}c}{} \\
\cmidrule(lr){3-4}
& & \textbf{Step} & \multicolumn{1}{cI}{\textbf{Latency}} & & & & & & & \multirow{-2}{*}{\textbf{Average ($\uparrow$)}} \\
\midrule
\textbf{Lumina-mGPT (7B)}     & 96.23s & $1.00\times$ & $1.00\times$ & 0.99 & 0.75 & 0.20 & 0.80 & 0.13 & 0.16 & 0.50 \\
\rowcolor{ourscol} + SSD (Ours)      & \textbf{13.09s} & $\mathbf{8.40\times}$   & $\mathbf{7.35\times}$   & 0.97 & 0.62 & 0.19 & 0.72 & 0.11 & 0.13 & \cellcolor{avgcolours}0.46 \\
\midrule
\textbf{Janus-Pro (7B)}       & 7.72s & $1.00\times$ & $1.00\times$ & 0.99 & 0.86 & 0.53 & 0.87 & 0.73 & 0.62 & 0.77 \\
\rowcolor{ourscol} + SSD (Ours)      & \textbf{1.45s} & $\mathbf{7.20\times}$   & $\mathbf{5.32\times}$   & 0.99 & 0.81 & 0.45 & 0.90 & 0.69 & 0.57 & \cellcolor{avgcolours}0.73 \\
\midrule
\textbf{Emu3 (8B)}            & 281.06s & $1.00\times$ & $1.00\times$ & 0.95 & 0.50 & 0.28 & 0.77 & 0.10 & 0.13 & 0.45 \\
\rowcolor{ourscol} + SSD (Ours)      & \textbf{25.00s} & $\mathbf{14.34\times}$   & $\mathbf{11.24\times}$   & 0.86 & 0.38 & 0.20 & 0.72 & 0.10 & 0.11 & \cellcolor{avgcolours}0.40 \\
\bottomrule
\end{tabular}%
}
\vspace{-2mm}
\end{table}

\begin{figure}[t!]
\centering
\begin{tikzpicture}
\begin{axis}[hide axis, xmin=0, xmax=1, ymin=0, ymax=1,
    width=\textwidth, height=2.2cm,
    legend columns=7,
    legend style={draw=none, font=\scriptsize, column sep=3pt,
        at={(0.5,0.5)}, anchor=center,
        /tikz/every even column/.append style={column sep=2pt}}]
\addlegendimage{bAR}      \addlegendentry{AR}
\addlegendimage{bSJD}     \addlegendentry{SJD}
\addlegendimage{bGSD}     \addlegendentry{GSD}
\addlegendimage{bLAN}     \addlegendentry{LANTERN}
\addlegendimage{bZIP}     \addlegendentry{ZipAR}
\addlegendimage{ssdcurve} \addlegendentry{\model{} (ours)}
\addlegendimage{ssdstarleg}  \addlegendentry{\model{} reported}
\end{axis}
\end{tikzpicture}

\vspace{-1mm}

\begin{tikzpicture}
\begin{axis}[ssdpanel, title={Janus-Pro-7B}, ylabel={DPG-Bench Average},
    xmin=0.3, xmax=8.3, ymin=80.9, ymax=84.9, ytick={81,82,83,84}, xtick={2,4,6,8}]
\addplot[ssdcurve]  coordinates {(3.42,84.44) (4.19,83.59) (5.70,83.40) (7.60,81.27)};
\addplot[bAR]       coordinates {(1.00,84.23)};
\addplot[bSJD]      coordinates {(1.50,84.01)};
\addplot[bGSD]      coordinates {(2.38,82.97)};
\addplot[bLAN]      coordinates {(0.84,84.07)};
\addplot[bZIP]      coordinates {(2.59,83.78)};
\addplot[ssdstar]   coordinates {(5.70,83.40)};
\end{axis}
\end{tikzpicture}%
\hspace{2pt}%
\begin{tikzpicture}
\begin{axis}[ssdpanel, title={Lumina-mGPT-7B}, ylabel={DPG-Bench Average},
    xmin=0.3, xmax=9.4, ymin=73.0, ymax=76.8, ytick={73,74,75,76}, xtick={2,4,6,8}]
\addplot[ssdcurve]  coordinates {(6.78,74.79) (7.35,74.57) (7.81,73.88) (8.65,73.75)};
\addplot[bAR]       coordinates {(1.00,76.30)};
\addplot[bSJD]      coordinates {(1.58,74.03)};
\addplot[bGSD]      coordinates {(2.52,73.37)};
\addplot[bLAN]      coordinates {(1.63,74.45)};
\addplot[bZIP]      coordinates {(2.64,74.62)};
\addplot[ssdstar]   coordinates {(7.35,74.57)};
\end{axis}
\end{tikzpicture}%
\hspace{2pt}%
\begin{tikzpicture}
\begin{axis}[ssdpanel, title={Emu3-8B}, ylabel={DPG-Bench Average},
    xmin=0.2, xmax=11.9, ymin=72.9, ymax=79.3, ytick={73,75,77,79}, xtick={2,4,6,8,10}]
\addplot[ssdcurve]  coordinates {(5.30,77.72) (6.26,77.33) (7.62,76.66) (11.03,76.03)};
\addplot[bAR]       coordinates {(1.00,78.69)};
\addplot[bSJD]      coordinates {(1.08,78.38)};
\addplot[bGSD]      coordinates {(2.90,77.57)};
\addplot[bLAN]      coordinates {(1.54,76.12)};
\addplot[bZIP]      coordinates {(3.54,73.36)};
\addplot[ssdstar]   coordinates {(11.03,76.03)};
\end{axis}
\end{tikzpicture}

\vspace{-2mm}
\caption{\textbf{SSD speed--quality trade-off curves.} Varying the verification budget controls the trade-off between generation quality and latency speedup. Each panel compares \model{} with decoding baselines using the same backbone as in Table~\ref{tab:dpgbench}. Yellow stars mark representative operating points that balance quality and speed, corresponding to the settings reported in Tables~\ref{tab:dpgbench} and~\ref{tab:geneval}. The top-right region is ideal; across backbones, \model{} consistently achieves comparable generation quality at substantially higher speedups.}
\label{fig:ssd_pareto}
\end{figure}
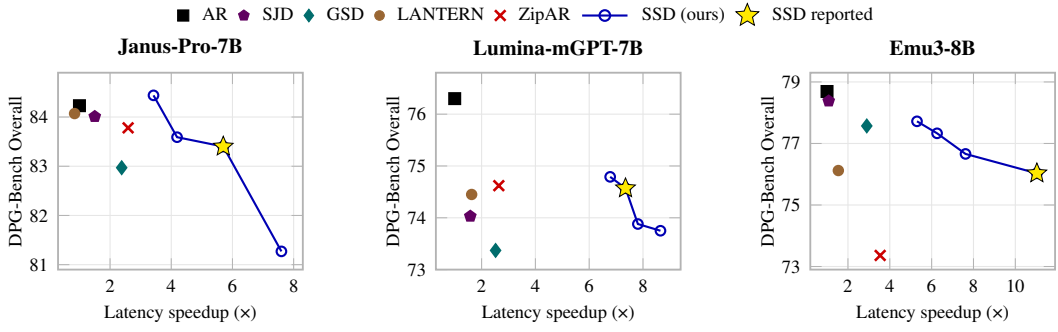

\section{Experiments}
\label{sec:experiment}

\paragraph{Models.}
We evaluate our method on three state-of-the-art autoregressive visual generation models: Janus-Pro-7B~\citep{chen2025januspro}, Lumina-mGPT-7B~\citep{liu2024lumina}, and Emu3-8B~\citep{wang2024emu3}, which produce token grids of $24{\times}24$ (576), $48{\times}48$ (2{,}304), and $90{\times}90$ (8{,}100 tokens), respectively. These sequence lengths span over an order of magnitude, providing a rigorous testbed to evaluate the generality of our approach. All pretrained backbones remain frozen; only the lightweight spatial draft predictors are trained.

\paragraph{Training.}
Each spatial draft predictor is trained via self-distillation~\citep{hinton2015distilling} using Midjourney prompts~\citep{yang2024midjourney}. To construct the training data, we first generate paired images for each input text prompt by performing inference with the original model. We then train the lightweight spatial MTP heads on those data. For number of training data, we use: 60{,}000 for Janus-Pro, 20{,}000 for Lumina-mGPT, and 5{,}000 for Emu3. Following the classifier-free guidance protocol~\citep{ho2022cfg}, we randomly drop 10\% of the text conditioning during training to enable unconditional generation. The five horizontal and vertical heads are trained independently using identical hyperparameters.

\paragraph{Metrics.}
To evaluate generation quality, we report results on two widely used text-to-image benchmarks. \textit{DPG-Bench}~\citep{hu2024ella} evaluates text-image semantic alignment under dense, compositionally rich prompts, providing a fine-grained assessment of how faithfully a model renders complex descriptions. \textit{GenEval}~\citep{ghosh2023geneval} measures compositional generation fidelity through object-focused evaluation, testing whether models correctly produce specified objects, attributes, and spatial layouts. For inference speed, we report efficiency across three key metrics that align with Table~\ref{tab:dpgbench}, \ref{tab:geneval}. Under \textbf{Acceleration ($\uparrow$) Step}, we report the reduction factor in the NFE of the backbone. Because the memory wall bottleneck is driven by the repeated loading of model parameters, reducing these forward steps directly minimizes memory I/O and yields actual time savings. We also report the absolute wall-clock \textbf{Latency ($\downarrow$)} in seconds, measured on an RTX Blackwell 6000 Pro GPU. Finally, \textbf{Acceleration ($\uparrow$) Latency} quantifies the relative wall-clock speedup achieved by our method compared to the baseline autoregressive approach.


\paragraph{Baselines.}
We compare against six decoding methods, all of which keep the pretrained backbone frozen. \textbf{AR} denotes standard autoregressive decoding with the unmodified pretrained model. \textbf{1D-MTP} uses only the horizontal heads from our framework with the same speculative decoding procedure and generation hyperparameters as the horizontal component of \model{}, isolating the contribution of vertical drafting. \textbf{SJD}~\citep{teng2024sjd} is a Jacobi-based parallel decoding method that iteratively refines multiple tokens along the 1D raster-scan sequence. \textbf{GSD}~\citep{so2025gsd} groups interchangeable visual tokens in the codebook and accepts a drafted token whenever it falls into the same group as the target token. \textbf{LANTERN}~\citep{jang2024lantern} relaxes the acceptance test using latent-space similarity between visual tokens, trading distributional exactness for higher acceptance rates. \textbf{ZipAR}~\citep{he2024zipar} is a training-free spatial scheme that decodes tokens from adjacent rows in parallel as soon as their local context is available.

\paragraph{\model{} Configuration.}
All three models share the same horizontal drafting setup: 5 tokens per row with 1 verification round. For vertical drafting, Lumina-mGPT drafts $k_v{=}2$ rows per iteration with staged verification---5 base rounds on both rows followed by 4 incremental rounds on the trailing row ($b{=}5,\;i{=}4$)---as its $48{\times}48$ grid benefits from multi-row drafting (Appendix~\ref{app:multi_row}). Janus-Pro and Emu3 draft a single row ($k_v{=}1$) with $r{=}2$ and $r{=}5$ verification rounds, respectively; on these grids the per-step cost of processing additional rows outweighs the step reduction from multi-row drafting.

\subsection{Main Results}

On DPG-Bench (Table~\ref{tab:dpgbench}), ZipAR is the fastest prior baseline on all three backbones, achieving $2.59\times$, $2.64\times$, and $3.54\times$ latency speedups on Janus-Pro, Lumina-mGPT, and Emu3, respectively. Our \model{} pushes these speedups to $5.70\times$, $7.35\times$, and $11.03\times$, while closely matching ZipAR's quality on Janus-Pro (83.40 vs.\ 83.78) and Lumina-mGPT (74.57 vs.\ 74.62), and improving it on Emu3 (76.03 vs.\ 73.36). The same operating points retain competitive quality on GenEval (Table~\ref{tab:geneval}). Figure~\ref{fig:ssd_pareto} further traces \model{}'s speed--quality frontier, showing that additional verification improves quality while retaining substantial speedups. Larger token grids provide more opportunities for row-parallel drafting and verification, leading to greater acceleration. Qualitative results are shown in Figure~\ref{fig:qualitative}, with additional baseline comparisons in Appendix~\ref{app:visualizations}.

\begin{figure}[t]
\centering
\vspace{-5mm}
\includegraphics[width=\textwidth,trim=18 0 20 0,clip]{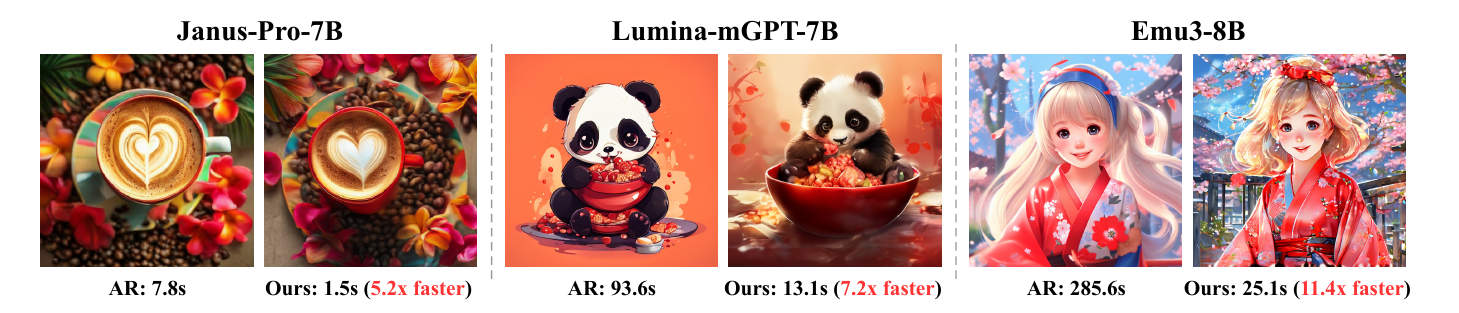}
\vspace{-4mm}
\caption{\textbf{Qualitative results.} Side-by-side comparison of AR baseline and \model{} across three models. Our method yields up to 11.03$\times$ speedup while preserving strong visual quality.}
\label{fig:qualitative}
\vspace{-2mm}
\end{figure}

\subsection{Analysis}
\label{sec:ablation}

\begin{table}[t]
\centering
\caption{To assess the importance of verification as auto correction, we compare three draft-update strategies on Lumina-mGPT-7B (\textbf{DPG-Bench}) under the fixed $b{=}5,\;i{=}4$ schedule. All methods start from the same SSD draft and use the same forward-step budget, differing only in how they verify and correct it. Argmax and direct sampling fail to preserve generation quality, while our acceptance-rejection as auto correction achieves substantially better quality.}
\vspace{2pt}
\label{tab:abl_accept}
\scriptsize
\begin{tabular}{@{}lIcIc@{}}
\toprule
\textbf{Method} & \textbf{Steps ($\downarrow$)} & \textbf{DPG-Bench Average ($\uparrow$)} \\
\midrule

(A) Greedy Selection (Argmax) & 280 & 66.41 \\
(B) Sampling & 280 & 66.16 \\
\rowcolor{ourscol} (C) Acceptance-rejection sampling (Ours) & 280 & \textbf{74.57} \\
\bottomrule
\end{tabular}
\vspace{-2mm}
\end{table}

\textbf{Importance of Verification as Auto Correction.} We compare three strategies for updating the same SSD draft on DPG-Bench with Lumina-mGPT-7B (Table~\ref{tab:abl_accept}): (1) taking the argmax of the backbone distribution produced by each forward step conditioned on the current draft, (2) sampling from the same distribution, and (3) our method, which uses a probability-based accept/reject step to decide whether to keep each drafted token or resample it. All strategies use the same number of backbone forward steps. Verification as auto correction yields substantially better generation quality.

\begin{figure}[t!]
\centering
\includegraphics[width=0.750\textwidth]{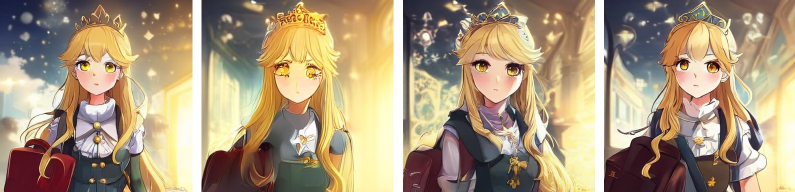}
\vspace{-2pt}
\makebox[\textwidth]{%
  \makebox[0.1875\textwidth]{\small AR}%
  \makebox[0.1875\textwidth]{\small $r=0$}%
  \makebox[0.1875\textwidth]{\small $r=1$}%
  \makebox[0.1875\textwidth]{\small $r=2$}%
}
\caption{Qualitative comparison of Janus-Pro-7B outputs under different vertical verification and correction rounds ($r$). The unverified draft ($r{=}0$) preserves coarse spatial layout, indicating that the spatial draft prediction acquires sufficient vertical coherence during training. Successive verification rounds progressively recover fine-grained detail and local coherence; by $r{=}2$ the output is visually indistinguishable from the autoregressive baseline.}
\label{fig:verify_samples}
\vspace{-2mm}
\end{figure}

\begin{table}[t]
\centering
\caption{Ablation on vertical verification rounds ($r$) on Janus-Pro-7B (\textbf{DPG-Bench}). $r$ denotes the number of verify-and-correct iterations applied to each vertically drafted row. The draft alone ($r{=}0$) produces coherent outputs, indicating that the spatial draft prediction captures sufficient spatial structure to serve as a reliable initialization. Subsequent verification rounds monotonically improve fidelity, recovering AR-level quality by $r{=}2$ while preserving substantial acceleration.}
\label{tab:verify_rounds}
\scriptsize
\begin{tabular}{@{}cIccIccIc@{}}
\toprule
\multirow{2}{*}{$r$} & \multirow{2}{*}{\textbf{Steps ($\downarrow$)}} & \multirow{2}{*}{\textbf{Latency ($\downarrow$)}} & \multicolumn{2}{cI}{\textbf{Acceleration ($\uparrow$)}} & \multirow{2}{*}{\textbf{DPG-Bench Average ($\uparrow$)}} \\
 & & & \textbf{Steps} & \textbf{Latency} & \\
\midrule
AR & 576 & 7.87s & $1.00\times$ & $1.00\times$ & 84.23 \\
\midrule
0 & 34 & \textbf{0.63s} & $\mathbf{16.90\times}$ & $\mathbf{12.50\times}$ & 71.65 \\
1 & 57 & 1.03s & $10.10\times$ & $7.60\times$ & 81.27 \\
\rowcolor{ourscol}
2 & 80 & 1.38s & $7.20\times$ & $5.70\times$ & \textbf{83.40} \\
\bottomrule
\end{tabular}
\end{table}

\textbf{The Effect of Number of Correct Rounds.} To examine the quality--speed tradeoff, we vary the number of per-row verification and correction rounds $r$ (Table~\ref{tab:verify_rounds}, Figure~\ref{fig:verify_samples}). The unverified draft ($r{=}0$) preserves coarse spatial layout but lacks fine detail. Each additional round progressively recovers local coherence, with quality improving sharply from $r{=}0$ (71.65) to $r{=}2$ (83.40). At $r{=}2$, the output is visually indistinguishable from the AR baseline while retaining $5.70\times$ latency acceleration.

\paragraph{Vertical Lookahead Depth of Multi-Row Anticipation.} To determine the optimal number of rows drafted per iteration, we vary the vertical lookahead depth $k_v$ with the staged schedule fixed at $b{=}5,\;i{=}4$ (Table~\ref{tab:abl_nvert} in Appendix~\ref{app:multi_row}). Increasing $k_v$ reduces the total step count, but with sharply diminishing returns: going from $k_v{=}1$ to $k_v{=}2$ removes 23 steps, whereas going from $k_v{=}2$ to $k_v{=}3$ removes only 8. Each forward pass meanwhile processes proportionally more tokens, so latency flattens out: the gap between $k_v{=}2$ and $k_v{=}3$ is $0.27$s, a tenth of the gap between $k_v{=}1$ and $k_v{=}2$. Quality, in contrast, peaks at $k_v{=}2$ (74.57 versus 73.42 and 74.31). We therefore use $k_v{=}2$ as the default, giving up $2\%$ latency relative to $k_v{=}3$ for the best generation quality.

\begin{table}[t!]
\begin{minipage}[t]{0.40\textwidth}
\vspace{0pt}
\centering
\scriptsize
\begin{tabular}{@{}cIc@{}}
\toprule
\textbf{Training samples} & \textbf{DPG-Bench ($\uparrow$)} \\
\midrule
5K & 54.67 \\
10K & 61.25 \\
20K & 64.20 \\
30K & 65.50 \\
\rowcolor{ourscol} 60K & \textbf{71.65} \\
\bottomrule
\end{tabular}
\end{minipage}%
\hfill
\begin{minipage}[t]{0.56\textwidth}
\vspace{0pt}
\captionof{table}{Effect of training data size on spatial draft prediction draft quality (Janus-Pro-7B, \textbf{DPG-Bench}, $r{=}0$). With only 5K samples the predictor already produces coherent drafts, and performance improves steadily through 60K. \colorbox{ourscol}{Blue} indicates our default.}
\label{tab:data_scaling}
\end{minipage}
\vspace{-4mm}
\end{table}

\textbf{Effect of Scaling Training Data.} To investigate the impact of training data scale, we vary the number of ground-truth samples used to train the spatial draft prediction (Table~\ref{tab:data_scaling}), evaluated at $r{=}0$ to isolate draft quality. The single-layer predictor already produces coherent drafts from just 5K hidden states (54.67), confirming that the continuous latent space provides a strong inductive bias for learning. Scaling from 5K to 60K yields a continuous upward trajectory, improving quality from 54.67 to 71.65. This upward trend shows no indication of saturation, suggesting that exposure to a broader distribution of visual representations continues to sharpen the predictor's spatial anticipation.

\section{Conclusion}

We introduce \modelfull{} (\model{}), an approach that leverages the intrinsic two-dimensional locality of images to overcome the memory wall in autoregressive visual inference. Our results demonstrate that this geometric shift substantially accelerates state-of-the-art generative models while maintaining strong generation quality. Our findings suggest that abandoning the linear constraints of language modeling to respect natural visual geometry is a promising direction for real-time, high-resolution generative models.

\subsubsection*{Acknowledgments}
This work used Purdue Anvil GPU through allocation
250774 from the Advanced Cyberinfrastructure Coordination Ecosystem: Services \& Support (ACCESS) program,
which is supported by U.S. National Science Foundation
grants \#2138259, \#2138286, \#2138307, \#2137603, and
\#2138296. This work used computing resources made available through the AMD University Program (AUP) AI \& HPC Cluster.

\bibliographystyle{iclr2027_conference}
\bibliography{references}


\appendix

\section{Technical appendices and supplementary material}
\label{app:main}

This supplementary material provides additional details that complement the main text. All experiments are carried out on NVIDIA RTX PRO 6000 Blackwell GPUs.

\subsection{Training and Inference Details}
\label{app:training}
\label{app:inference}

See Table~\ref{tab:app_config}.

\begin{table}[H]
\centering
\caption{Training and inference configuration for the spatial drafting heads. The learning-rate schedule and the optimizer are identical for all three backbones.}
\label{tab:app_config}
\scriptsize
\begin{tabular}{@{}l@{\hspace{1.5em}}Iccc@{}}
\toprule
& \textbf{Janus-Pro-7B} & \textbf{Lumina-mGPT-7B} & \textbf{Emu3-8B} \\
\midrule
Token grid              & $24{\times}24$ & $48{\times}48$ & $90{\times}90$ \\
Params per head (M)     & 134.2 & 134.2 & 134.2 \\
\midrule
\multicolumn{4}{@{}l}{\emph{Training}} \\
\quad Training prompts        & 60{,}000 & 20{,}000 & 5{,}000 \\
\quad Batch size / GPU        & 128 & 32 & 8 \\
\quad GPUs                    & 4 & 4 & 4 \\
\quad Epochs                  & 3 & 3 & 3 \\
\quad Gradient accumulation   & 1 & 1 & 1 \\
\quad CFG drop rate           & 10\% & 10\% & 10\% \\
\quad Learning rate           & \multicolumn{3}{c}{$1{\times}10^{-4}\!\rightarrow\!1{\times}10^{-5}$, 20 warmup steps} \\
\quad Optimizer               & \multicolumn{3}{c}{AdamW ($\beta_1{=}0.9$, $\beta_2{=}0.95$, wd${=}0.01$)} \\
\quad Training cost (GPU-h)   & 3.0 & 5.6 & 3.3 \\
\midrule
\multicolumn{4}{@{}l}{\emph{Inference}} \\
\quad Temperature             & 1.0 & 1.0 & 1.0 \\
\quad CFG weight              & 5.0 & 4.0 & 3.0 \\
\quad Horizontal heads $k_h$  & 5 & 5 & 5 \\
\quad Horizontal verify rounds & 1 & 1 & 1 \\
\quad Vertical heads $k_v$    & 1 & 2 & 1 \\
\quad Vertical verify rounds  & $r{=}2$ & $b{=}5,\;i{=}4$ & $r{=}5$ \\
\bottomrule
\end{tabular}
\end{table}

\subsection{Speed--Quality Trade-off}
\label{app:pareto}

The number of verification rounds provides an explicit control over the speed--quality trade-off. Table~\ref{tab:app_pareto} sweeps this budget across all three backbones, with ZipAR as the fastest baseline. Increasing the verification budget consistently improves quality, with diminishing returns at larger budgets: Janus-Pro reaches 84.44 at $r{=}4$, slightly above AR (84.23); Emu3 improves from 76.03 at $r{=}5$ to 77.72 at $r{=}12$, approaching AR (78.69); and Lumina-mGPT gains little beyond $i{=}4$. Across all three backbones, \model{} provides an operating point that is both faster and higher-quality than ZipAR while remaining faster throughout the sweep.

\begin{table}[t]
\centering
\caption{Speed--quality trade-off on \textbf{DPG-Bench} as a function of the verification budget. For Lumina-mGPT the schedule uses $b$ verification rounds on the first row of each block and $i$ further rounds on each subsequent row. ZipAR is the fastest baseline on all three backbones. \colorbox{ourscol}{Blue} rows indicate the default configuration reported in Table~\ref{tab:dpgbench}.}
\label{tab:app_pareto}
\scriptsize
\begin{tabular}{@{}lIccIccIc@{}}
\toprule
\multirow{2}{*}{\textbf{Method}} & \multirow{2}{*}{\textbf{Steps ($\downarrow$)}} & \multirow{2}{*}{\textbf{Latency ($\downarrow$)}} & \multicolumn{2}{cI}{\textbf{Acceleration ($\uparrow$)}} & \multirow{2}{*}{\textbf{DPG-Bench Average ($\uparrow$)}} \\
\cmidrule(lr){4-5}
& & & \textbf{Step} & \textbf{Latency} & \\
\midrule
\multicolumn{6}{@{}l}{\textbf{Janus-Pro-7B}} \\
AR (baseline)       & 576 & 7.87s & $1.00\times$ & $1.00\times$ & 84.23 \\
ZipAR ($w{=}8$)     & 209 & 3.04s & $2.76\times$ & $2.59\times$ & 83.78 \\
\model{} ($r{=}0$)  & 34 & 0.63s & $16.90\times$ & $12.50\times$ & 71.65 \\
\model{} ($r{=}1$)  & 57 & 1.03s & $10.10\times$ & $7.60\times$ & 81.27 \\
\rowcolor{ourscol} \model{} ($r{=}2$) & 80 & \textbf{1.38s} & $\mathbf{7.20\times}$ & $\mathbf{5.70\times}$ & 83.40 \\
\model{} ($r{=}3$)  & 103 & 1.88s & $5.59\times$ & $4.19\times$ & 83.59 \\
\model{} ($r{=}4$)  & 126 & 2.30s & $4.57\times$ & $3.42\times$ & \textbf{84.44} \\
\midrule
\multicolumn{6}{@{}l}{\textbf{Lumina-mGPT-7B}} \\
AR (baseline)       & 2{,}352 & 96.23s & $1.00\times$ & $1.00\times$ & 76.30 \\
ZipAR ($w{=}16$)    & 801 & 36.45s & $2.94\times$ & $2.64\times$ & 74.62 \\
\model{} ($b{=}5,\,i{=}1$) & 211 & \textbf{9.97s} & $\mathbf{11.15\times}$ & $\mathbf{9.65\times}$ & 72.22 \\
\model{} ($b{=}5,\,i{=}2$) & 234 & 11.12s & $10.05\times$ & $8.65\times$ & 73.75 \\
\model{} ($b{=}5,\,i{=}3$) & 257 & 12.32s & $9.15\times$ & $7.81\times$ & 73.88 \\
\rowcolor{ourscol} \model{} ($b{=}5,\,i{=}4$) & 280 & 13.09s & $8.40\times$ & $7.35\times$ & 74.57 \\
\model{} ($b{=}5,\,i{=}5$) & 303 & 14.19s & $7.76\times$ & $6.78\times$ & \textbf{74.79} \\
\midrule
\multicolumn{6}{@{}l}{\textbf{Emu3-8B}} \\
AR (baseline)       & 8{,}190 & 281.87s & $1.00\times$ & $1.00\times$ & 78.69 \\
ZipAR ($w{=}16$)    & 1{,}515 & 79.71s & $5.41\times$ & $3.54\times$ & 73.36 \\
\rowcolor{ourscol} \model{} ($r{=}5$)  & 571 & \textbf{25.55s} & $\mathbf{14.34\times}$ & $\mathbf{11.03\times}$ & 76.03 \\
\model{} ($r{=}8$)  & 838 & 37.00s & $9.77\times$ & $7.62\times$ & 76.66 \\
\model{} ($r{=}10$) & 1{,}016 & 44.99s & $8.06\times$ & $6.26\times$ & 77.33 \\
\model{} ($r{=}12$) & 1{,}194 & 53.20s & $6.86\times$ & $5.30\times$ & \textbf{77.72} \\
\bottomrule
\end{tabular}
\end{table}

\subsection{Perceptual Fidelity}
\label{app:fid}

DPG-Bench and GenEval primarily measure text--image alignment rather than perceptual quality. We therefore additionally evaluate FID on COCO2017-val 5K with Janus-Pro-7B (Table~\ref{tab:app_fid}). Verification rapidly closes the gap from the unverified draft: at $r{=}2$, \model{} reaches 28.95 FID while retaining a $5.70\times$ latency speedup. Over five random seeds (Table~\ref{tab:app_fid_seeds}), \model{} obtains a clean-FID of $28.89 \pm 0.24$ versus $29.13 \pm 0.08$ for AR. The difference of $0.25$ has a 95\% confidence interval of $[-0.04, +0.54]$ (Welch's $p{=}0.08$), indicating no statistically significant difference in FID.

\begin{table}[t]
\centering
\caption{FID on COCO2017-val 5K (Janus-Pro-7B) as a function of verification rounds $r$. \colorbox{ourscol}{Blue} indicates our default configuration.}
\label{tab:app_fid}
\scriptsize
\begin{tabular}{@{}lIcIcc@{}}
\toprule
\textbf{Method} & \textbf{FID ($\downarrow$)} & \textbf{Latency ($\downarrow$)} & \textbf{Acceleration ($\uparrow$)} \\
\midrule
AR (baseline)      & 29.19 & 7.87s & $1.00\times$ \\
\midrule
\model{} ($r{=}0$) & 59.19 & \textbf{0.63s} & $\mathbf{12.49\times}$ \\
\model{} ($r{=}1$) & 31.26 & 1.03s & $7.64\times$ \\
\rowcolor{ourscol} \model{} ($r{=}2$) & 28.95 & 1.38s & $5.70\times$ \\
\model{} ($r{=}3$) & \textbf{28.70} & 1.85s & $4.25\times$ \\
\bottomrule
\end{tabular}
\end{table}

\begin{table}[t]
\centering
\caption{clean-FID on COCO2017-val 5K (Janus-Pro-7B) over five random seeds, reported as mean $\pm$ s.d. The difference in favour of \model{} is $+0.25$ FID with a 95\% confidence interval of $[-0.04, +0.54]$ (Welch's two-sample $t$-test, $p{=}0.08$), i.e.\ the two are statistically indistinguishable.}
\label{tab:app_fid_seeds}
\scriptsize
\begin{tabular}{@{}lIcIcc@{}}
\toprule
\textbf{Method} & \textbf{clean-FID ($\downarrow$)} & \textbf{Latency ($\downarrow$)} & \textbf{Acceleration ($\uparrow$)} \\
\midrule
AR (baseline) & $29.13 \pm 0.08$ & 7.87s & $1.00\times$ \\
\rowcolor{ourscol} \model{} ($r{=}2$) & $\mathbf{28.89 \pm 0.24}$ & \textbf{1.38s} & $\mathbf{5.70\times}$ \\
\bottomrule
\end{tabular}
\end{table}

\subsection{Human Evaluation}
\label{app:human_eval}

\begin{figure}[t]
\centering
\includegraphics[width=0.42\textwidth,trim=100 14 55 13,clip]{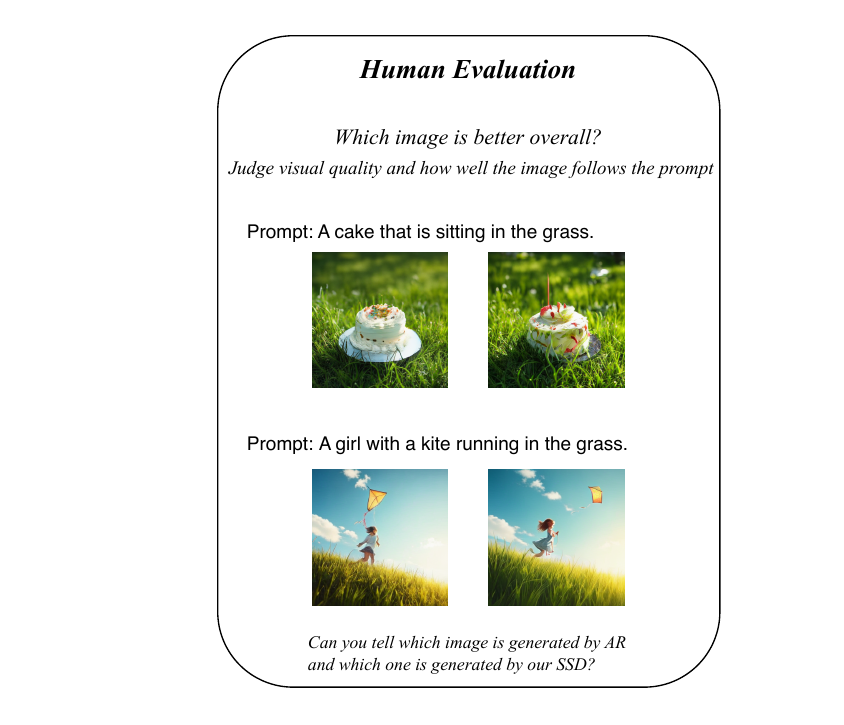}
\caption{Two samples of the human preference study. Each screen shows a COCO2017-val prompt together with the AR and \model{} ($r{=}2$) images generated from it under the same seed, placed side by side in randomized order and without method labels; the evaluator selects the image that is better overall in visual quality and prompt adherence.}
\label{fig:app_human}
\end{figure}

Clean-FID compares distributions rather than individual images, and therefore cannot tell whether a human observer can distinguish a given \model{} sample from its AR counterpart. We therefore conduct a pairwise preference study on 50 randomly sampled COCO2017-val prompts. For each prompt, Janus-Pro-7B generates one image with AR decoding and one with \model{} ($r{=}2$) under the same seed, and seven evaluators compare the pair in randomized order without method labels; Figure~\ref{fig:app_human} shows the evaluation interface and two example pairs. As shown in Table~\ref{tab:app_human}, AR and \model{} receive preference rates of $54.9 \pm 7.0\%$ and $45.1 \pm 7.0\%$, respectively. Evaluator-level variation spans the $50\%$ indifference point, indicating no clear perceptual preference between the two methods, while \model{} generates images $5.70\times$ faster.

\begin{table}[t]
\centering
\caption{Human pairwise preference on 50 COCO2017-val prompts (Janus-Pro-7B), reported as mean $\pm$ s.d.\ across seven evaluators.}
\label{tab:app_human}
\scriptsize
\begin{tabular}{@{}lIcIc@{}}
\toprule
\textbf{Method} & \textbf{Win rate ($\uparrow$)} & \textbf{Latency speedup ($\uparrow$)} \\
\midrule
AR (baseline) & $54.9 \pm 7.0\%$ & $1.00\times$ \\
\rowcolor{ourscol} \model{} ($r{=}2$) & $45.1 \pm 7.0\%$ & $\mathbf{5.70\times}$ \\
\bottomrule
\end{tabular}
\end{table}

\subsection{Characterizing the Deviation from Autoregressive Sampling}
\label{app:bias}

\model{} verifies each row in parallel using a context that may still contain drafted tokens, it does not preserve the exact autoregressive distribution. We quantify this deviation on Janus-Pro-7B using $K{=}100$ randomly sampled COCO2017-val prompts and $N{=}10$ random seeds per prompt for both AR and \model{} ($r{=}2$). For each prompt, we estimate the mean $\mu$ and covariance $\Sigma$ of the generated images in Janus-Pro's visual feature space and pool the resulting statistics across prompts (Table~\ref{tab:app_bias}).

The squared distance between the AR and \model{} centers is $0.0212$, corresponding to $0.70$ of the AR conditional standard deviation, while the feature-space spread increases by $25\%$ ($\mathrm{tr}\,\Sigma_{\mathrm{AR}} = 0.0428$ versus $\mathrm{tr}\,\Sigma_{\mathrm{SSD}} = 0.0533$). Thus, the center shift induced by parallel verification is small relative to the natural seed-to-seed variability of AR generation.

\begin{table}[t]
\centering
\caption{Deviation of \model{} ($r{=}2$) from AR generation on Janus-Pro-7B, measured in Janus-Pro's visual feature space over $K{=}100$ COCO2017-val prompts $\times$ $N{=}10$ seeds.}
\label{tab:app_bias}
\scriptsize
\begin{tabular}{@{}lIcIc@{}}
\toprule
\textbf{Effect} & \textbf{Measured} & \textbf{Relative to AR} \\
\midrule
Location (center shift) & $\lVert \mu_{\mathrm{SSD}} - \mu_{\mathrm{AR}} \rVert^2 = 0.0212$ & $0.70$ AR cond.\ s.d. \\
Scale (spread) & $\mathrm{tr}\,\Sigma_{\mathrm{AR}} = 0.0428$,\; $\mathrm{tr}\,\Sigma_{\mathrm{SSD}} = 0.0533$ & $+25\%$ \\
\bottomrule
\end{tabular}
\end{table}

\subsection{Computational Cost of the Drafting Heads}
\label{app:cost}

We quantify the overhead of the drafting heads at the operating points reported in Table~\ref{tab:dpgbench}. Each head contains $134.2$\,M parameters (Table~\ref{tab:app_config}), but only a subset is active at each stage. The five horizontal heads add $0.671$\,B parameters and $1.34$\,GB of bf16 memory but are used only to complete the first row of each block. During the dominant vertical stage, \model{} activates only one or two vertical heads, adding $0.134$--$0.268$\,B parameters ($1.6$--$3.8\%$ of the backbone), $0.27$--$0.54$\,GB of memory, and $1.5$--$2.0\%$ per-step FLOPs overhead. Training the heads requires only $3.0$--$5.6$ GPU-hours.

\model{} reduces sequential forward evaluations rather than total arithmetic. Table~\ref{tab:app_flops} accounts for all backbone and verification passes, attention, drafting heads, activations, and KV-cache memory. Compared with AR, \model{} performs $2.7$--$7.7\times$ more total FLOPs and uses $2.4$--$4.3$\,GB more peak memory, yet achieves $5.70$--$11.03\times$ lower wall-clock latency. This behavior is consistent with memory-bandwidth-bound decoding, where repeated backbone parameter loading dominates latency.

\begin{table}[t]
\centering
\caption{End-to-end compute and memory per generated image, measured under identical optimized inference loops. Drafting-head rows are split into the horizontal heads, which run only on the first row of each block, and the vertical heads, which run on all remaining rows.}
\label{tab:app_flops}
\scriptsize
\begin{tabular}{@{}lIccIccIcc@{}}
\toprule
& \multicolumn{2}{cI}{\textbf{Janus-Pro-7B}} & \multicolumn{2}{cI}{\textbf{Lumina-mGPT-7B}} & \multicolumn{2}{c}{\textbf{Emu3-8B}} \\
\cmidrule(lr){2-3}\cmidrule(lr){4-5}\cmidrule(lr){6-7}
& \textbf{AR} & \textbf{\model{} ($r{=}2$)} & \textbf{AR} & \textbf{\model{} ($b{=}5,i{=}4$)} & \textbf{AR} & \textbf{\model{} ($r{=}5$)} \\
\midrule
\multicolumn{7}{@{}l}{\emph{Compute per image} (TFLOPs)} \\
\quad Backbone, all forward passes  & 16.6 & 43.8 & 65.5 & 499 & 255 & 1{,}521 \\
\quad Attention                     & 0.224 & 0.646 & 3.08 & 24.3 & 35.9 & 216 \\
\quad Drafting heads, horizontal    & --- & 0.012 & --- & 0.025 & --- & 0.048 \\
\quad Drafting heads, vertical      & --- & 0.296 & --- & 2.42 & --- & 4.30 \\
\quad \textbf{End-to-end}           & \textbf{16.8} & \textbf{44.8} & \textbf{68.6} & \textbf{526} & \textbf{291} & \textbf{1{,}742} \\
\midrule
\multicolumn{7}{@{}l}{\emph{Memory} (GB)} \\
\quad Model weights                 & 14.86 & 16.47 & 14.37 & 16.25 & 18.23 & 19.84 \\
\quad\quad \emph{of which drafting heads} & --- & 1.61 & --- & 1.88 & --- & 1.61 \\
\quad Activations $+$ KV cache      & 1.33 & 2.12 & 2.61 & 2.79 & 4.36 & 7.02 \\
\quad \textbf{Peak GPU memory}      & \textbf{16.18} & \textbf{18.59} & \textbf{16.98} & \textbf{19.04} & \textbf{22.59} & \textbf{26.86} \\
\midrule
Wall-clock latency (s)              & 7.87 & \textbf{1.38} & 96.23 & \textbf{13.09} & 281.87 & \textbf{25.55} \\
\bottomrule
\end{tabular}
\end{table}

Figure~\ref{fig:draft_cost} reports drafting cost, while Figure~\ref{fig:mtp_analysis} shows the near-linear token acceleration enabled by parallel verification on Lumina-mGPT. Table~\ref{tab:app_cost_backbones} extends the normalized drafting and verification cost measurements to all three backbones. At the one-row operating point, drafting costs only $0.063$, $0.040$, and $0.098$ AR steps on Janus-Pro, Lumina-mGPT, and Emu3, respectively, while verification costs $1.26$, $1.20$, and $1.22$ AR steps. Generating the same rows autoregressively requires $24$, $48$, and $90$ sequential steps. Thus, a row-level evaluation is only modestly more expensive than a single-token evaluation while replacing many sequential forward evaluations.

\begin{table}[t]
\centering
\caption{Drafting and verification cost across all three backbones, normalized by one AR decoding step at the same generation position. $K$ is a multiple of the grid width because drafting is row-parallel. Values are means over four generation positions with 50 timed repetitions each.}
\label{tab:app_cost_backbones}
\scriptsize
\begin{tabular}{@{}lIcIcc@{}}
\toprule
\textbf{Model (grid)} & \textbf{Tokens $K$} & \textbf{Drafting (AR steps)} & \textbf{Verification (AR steps)} \\
\midrule
\multirow{5}{*}{Janus-Pro-7B ($24{\times}24$)}
  & 24  & 0.063 & 1.256 \\
  & 48  & 0.091 & 1.263 \\
  & 72  & 0.142 & 1.365 \\
  & 96  & 0.155 & 1.407 \\
  & 120 & 0.174 & 1.433 \\
\midrule
\multirow{5}{*}{Lumina-mGPT-7B ($48{\times}48$)}
  & 48  & 0.040 & 1.202 \\
  & 96  & 0.055 & 1.220 \\
  & 144 & 0.078 & 1.274 \\
  & 192 & 0.081 & 1.300 \\
  & 240 & 0.091 & 1.335 \\
\midrule
\multirow{5}{*}{Emu3-8B ($90{\times}90$)}
  & 90  & 0.098 & 1.218 \\
  & 180 & 0.150 & 1.525 \\
  & 270 & 0.236 & 2.071 \\
  & 360 & 0.271 & 2.506 \\
  & 450 & 0.337 & 2.571 \\
\bottomrule
\end{tabular}
\end{table}

\subsection{Configuration Selection}
\label{app:config_selection}

The configurations reported in Table~\ref{tab:dpgbench} are selected on 100 held-out Midjourney prompts. Tables~\ref{tab:app_sel_janus}--\ref{tab:app_sel_emu3} report the full sweeps. For each backbone, we sweep the vertical lookahead $v$, base verification budget $b$, and incremental budget $i$, and select an reasonable point that balances acceptance rate and inference latency while visually inspecting samples at each candidate setting. 

\begin{table}[t]
\centering
\caption{Configuration sweep on Janus-Pro-7B over 100 held-out Midjourney prompts. AR reference: 576 steps, 7.87\,s. Accept rate is undefined at $b{=}0$ because no verification is performed. \colorbox{ourscol}{Blue} indicates the deployed configuration.}
\label{tab:app_sel_janus}
\scriptsize
\begin{tabular}{@{}lIcccIcccIc@{}}
\toprule
\textbf{Config} & $v$ & $b$ & $i$ & \textbf{Steps ($\downarrow$)} & \textbf{Latency ($\downarrow$)} & \textbf{Speedup ($\uparrow$)} & \textbf{Accept rate ($\uparrow$)} \\
\midrule
v1b0i0 & 1 & 0 & 0 & 34 & 0.59s & $13.42\times$ & --- \\
v1b1i0 & 1 & 1 & 0 & 57 & 0.98s & $8.00\times$ & 0.1657 \\
\rowcolor{ourscol} v1b2i0 ($r{=}2$) & 1 & 2 & 0 & 80 & 1.38s & $5.70\times$ & 0.4305 \\
v1b3i0 & 1 & 3 & 0 & 103 & 1.88s & $4.19\times$ & 0.5571 \\
v1b4i0 & 1 & 4 & 0 & 126 & 2.30s & $3.42\times$ & 0.6302 \\
v1b5i0 & 1 & 5 & 0 & 149 & 2.77s & $2.84\times$ & 0.6778 \\
\midrule
v2b2i1 & 2 & 2 & 1 & 69 & 1.19s & $6.61\times$ & 0.3378 \\
v2b2i2 & 2 & 2 & 2 & 80 & 1.39s & $5.65\times$ & 0.4418 \\
\midrule
v3b2i1 & 3 & 2 & 1 & 65 & 1.12s & $7.02\times$ & 0.2978 \\
v3b2i2 & 3 & 2 & 2 & 80 & 1.39s & $5.65\times$ & 0.4471 \\
\bottomrule
\end{tabular}
\end{table}

\begin{table}[t]
\centering
\caption{Configuration sweep on Lumina-mGPT-7B over 100 held-out Midjourney prompts. AR reference: 2{,}352 steps, 96.23\,s. \colorbox{ourscol}{Blue} indicates the deployed configuration.}
\label{tab:app_sel_lumina}
\scriptsize
\begin{tabular}{@{}lIcccIcccIc@{}}
\toprule
\textbf{Config} & $v$ & $b$ & $i$ & \textbf{Steps ($\downarrow$)} & \textbf{Latency ($\downarrow$)} & \textbf{Speedup ($\uparrow$)} & \textbf{Accept rate ($\uparrow$)} \\
\midrule
v1b0i0 & 1 & 0 & 0 & 68  & 3.21s  & $29.95\times$ & --- \\
v1b1i0 & 1 & 1 & 0 & 115 & 5.43s  & $17.71\times$ & 0.6894 \\
v1b2i0 & 1 & 2 & 0 & 162 & 7.65s  & $12.57\times$ & 0.6636 \\
v1b3i0 & 1 & 3 & 0 & 209 & 9.88s  & $9.74\times$  & 0.7027 \\
v1b4i0 & 1 & 4 & 0 & 256 & 12.09s & $7.96\times$  & 0.7364 \\
v1b5i0 & 1 & 5 & 0 & 303 & 14.22s & $6.77\times$  & 0.7605 \\
v1b6i0 & 1 & 6 & 0 & 350 & 16.35s & $5.89\times$  & 0.7846 \\
v1b7i0 & 1 & 7 & 0 & 397 & 18.47s & $5.21\times$  & 0.7970 \\
v1b8i0 & 1 & 8 & 0 & 444 & 20.60s & $4.67\times$  & 0.8069 \\
\midrule
v2b5i1 & 2 & 5 & 1 & 211 & 9.97s  & $9.65\times$  & 0.6758 \\
v2b5i2 & 2 & 5 & 2 & 234 & 11.12s & $8.65\times$  & 0.7159 \\
v2b5i3 & 2 & 5 & 3 & 257 & 12.32s & $7.81\times$  & 0.7441 \\
\rowcolor{ourscol} v2b5i4 ($b{=}5,i{=}4$) & 2 & 5 & 4 & 280 & 13.09s & $7.35\times$ & 0.7645 \\
v2b5i5 & 2 & 5 & 5 & 303 & 14.19s & $6.78\times$  & 0.7745 \\
\midrule
v3b5i1 & 3 & 5 & 1 & 179 & 8.46s  & $11.38\times$ & 0.6263 \\
v3b5i2 & 3 & 5 & 2 & 210 & 9.92s  & $9.70\times$  & 0.6979 \\
v3b5i3 & 3 & 5 & 3 & 241 & 11.41s & $8.43\times$  & 0.7369 \\
v3b5i4 & 3 & 5 & 4 & 272 & 12.82s & $7.51\times$  & 0.7605 \\
v3b5i5 & 3 & 5 & 5 & 303 & 14.22s & $6.77\times$  & 0.7771 \\
\bottomrule
\end{tabular}
\end{table}

\begin{table}[t]
\centering
\caption{Configuration sweep on Emu3-8B over 100 held-out Midjourney prompts. AR reference: 8{,}190 steps, 278.33\,s on the same held-out set. \colorbox{ourscol}{Blue} indicates the deployed configuration. \textsuperscript{*}Linearly interpolated: this configuration had no contention-free measurement window.}
\label{tab:app_sel_emu3}
\scriptsize
\begin{tabular}{@{}lIcccIcccIc@{}}
\toprule
\textbf{Config} & $v$ & $b$ & $i$ & \textbf{Steps ($\downarrow$)} & \textbf{Latency ($\downarrow$)} & \textbf{Speedup ($\uparrow$)} & \textbf{Accept rate ($\uparrow$)} \\
\midrule
v1b0i0 & 1 & 0 & 0 & 126 & 5.30s  & $52.5\times$  & --- \\
v1b1i0 & 1 & 1 & 0 & 215 & 9.18s  & $30.3\times$  & 0.2067 \\
v1b2i0 & 1 & 2 & 0 & 304 & 13.07s & $21.3\times$  & 0.4913 \\
v1b3i0 & 1 & 3 & 0 & 393 & 16.94s & $16.4\times$  & 0.6585 \\
v1b4i0 & 1 & 4 & 0 & 482 & 20.85s & $13.4\times$  & 0.7212 \\
\rowcolor{ourscol} v1b5i0 ($r{=}5$) & 1 & 5 & 0 & 571 & 24.71s & $11.3\times$ & 0.7512 \\
v1b6i0 & 1 & 6 & 0 & 660 & 28.57s & $9.74\times$ & 0.7739 \\
v1b7i0 & 1 & 7 & 0 & 749 & 32.46s & $8.58\times$  & 0.7906 \\
v1b8i0 & 1 & 8 & 0 & 838 & 36.35s & $7.66\times$  & 0.7991 \\
\midrule
v2b6i1 & 2 & 6 & 1 & 440 & 22.03s & $12.63\times$ & 0.6958 \\
v2b6i2 & 2 & 6 & 2 & 484 & 23.98s & $11.61\times$ & 0.7392 \\
v2b6i3 & 2 & 6 & 3 & 528 & 25.90s & $10.75\times$ & 0.7634 \\
v2b6i4 & 2 & 6 & 4 & 572 & 27.84s & $10.00\times$ & 0.7769 \\
v2b6i5 & 2 & 6 & 5 & 616 & 29.75s & $9.35\times$  & 0.7856 \\
v2b6i6 & 2 & 6 & 6 & 660 & 31.68s & $8.78\times$  & 0.7950 \\
\midrule
v3b6i1 & 3 & 6 & 1 & 365 & 21.56s & $12.91\times$ & 0.6606 \\
v3b6i2 & 3 & 6 & 2 & 424 & 24.50s & $11.36\times$ & 0.7205 \\
v3b6i3 & 3 & 6 & 3 & 483 & 27.41s & $10.16\times$ & 0.7573 \\
v3b6i4 & 3 & 6 & 4 & 542 & 30.33s\textsuperscript{*} & $9.18\times$ & 0.7804 \\
v3b6i5 & 3 & 6 & 5 & 601 & 33.27s & $8.37\times$  & 0.7935 \\
v3b6i6 & 3 & 6 & 6 & 660 & 36.16s & $7.70\times$  & 0.8025 \\
\bottomrule
\end{tabular}
\end{table}

\subsection{Generalization of the Drafting Heads}
\label{app:generalization}

The drafting heads are trained on Midjourney prompts but evaluated on DPG-Bench and GenEval, which differ substantially in length, vocabulary, style, and compositional structure (Table~\ref{tab:app_prompt_shift}). Roughly half of the DPG-Bench word types are unseen during training, the corpora share almost no 5-grams, and the evaluation benchmarks place substantially greater emphasis on spatial relations, attribute binding, and counting. Despite this shift, Emu3's heads, trained on only 5{,}000 prompts, reach 76.03 on DPG-Bench compared with 78.69 for AR while providing an $11.03\times$ wall-clock speedup. Table~\ref{tab:app_backbones} further shows that the same head architecture and training recipe transfer across three backbones with different resolutions, visual tokenizers, and a $14\times$ range in tokens per image.

\begin{table}[t]
\centering
\caption{Training prompts (Midjourney) versus evaluation benchmarks. Vocabulary and 5-gram statistics are computed against Emu3's 5{,}000 training prompts, the smallest of the three training sets.}
\label{tab:app_prompt_shift}
\scriptsize
\begin{tabular}{@{}lIccc@{}}
\toprule
& \textbf{Training (Midjourney)} & \textbf{DPG-Bench} & \textbf{GenEval} \\
\midrule
Number of prompts             & 5K--60K & 1{,}065 & 553 \\
Median length (words)         & 14      & 65      & 8 \\
Exact prompt overlap          & ---     & 0       & 0 \\
Word types unseen in training & ---     & 49.4\%  & 29.3\% \\
5-grams shared with training  & ---     & 0.35\%  & 0.00\% \\
Midjourney style modifiers    & 18.0\%  & 7.1\%   & 0.0\% \\
Explicit spatial relations    & 4.5\%   & 39.6\%  & 18.1\% \\
Colour--attribute binding     & 31.2\%  & 72.0\%  & 35.1\% \\
Explicit counting             & 4.1\%   & 5.6\%   & 11.4\% \\
\bottomrule
\end{tabular}
\end{table}

\begin{table}[t]
\centering
\caption{The same head architecture and training recipe are applied to three backbones spanning different resolutions, visual tokenizers, and a $14\times$ range in tokens per image.}
\label{tab:app_backbones}
\scriptsize
\begin{tabular}{@{}lIccc@{}}
\toprule
& \textbf{Janus-Pro-7B} & \textbf{Lumina-mGPT-7B} & \textbf{Emu3-8B} \\
\midrule
Resolution        & $384^2$ & $768^2$ & $720^2$ \\
Visual tokenizer  & VQ-16 & Chameleon VQ & Emu3 VQ \\
Codebook size     & 16{,}384 & 8{,}192 & 32{,}768 \\
Token grid        & $24{\times}24$ & $48{\times}48$ & $90{\times}90$ \\
Tokens per image  & 576 & 2{,}304 & 8{,}100 \\
Training images   & 60{,}000 & 20{,}000 & 5{,}000 \\
DPG-Bench, AR $\to$ \model{} & 84.23 $\to$ 83.40 & 76.30 $\to$ 74.57 & 78.69 $\to$ 76.03 \\
Wall-clock speedup & $5.70\times$ & $7.35\times$ & $11.03\times$ \\
\bottomrule
\end{tabular}
\end{table}

\subsection{Prediction Target Space}
\label{app:target_space}

\begin{figure}[t]
\begin{minipage}[t]{0.325\textwidth}
\vspace{-1pt}
\centering
\includegraphics[width=\textwidth,trim=18 25 20 12,clip]{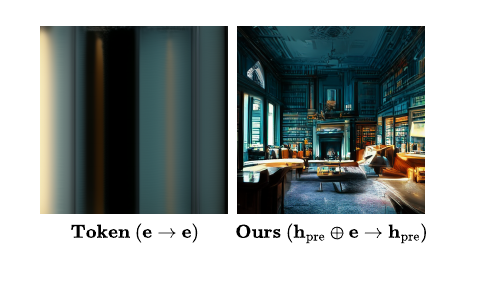}
\end{minipage}%
\hfill
\begin{minipage}[t]{0.66\textwidth}
\vspace{1pt}
\captionof{figure}{Samples generated by Janus-Pro-7B with different spatial draft prediction configurations at $r{=}0$ (no verification). The token-space head ($\mathbf{e} \!\to\! \mathbf{e}$) fails to reconstruct meaningful visual content, whereas our hidden-state head ($\mathbf{h}_\text{pre}$$\oplus$$\mathbf{e} \!\to\! \mathbf{h}_\text{pre}$) generates coherent images. Highlighting the importance of predicting the right target in our SSD.}
\label{fig:abl_design_samples}
\end{minipage}
\vspace{-5mm}
\end{figure}

\begin{table}[t]
\begin{minipage}[c]{0.36\textwidth}
\centering
\scriptsize
\begin{tabular}{@{}llIc@{}}
\toprule
\textbf{Input} & \textbf{Target} & \textbf{DPG-Bench Average ($\uparrow$)} \\
\midrule
$\mathbf{e}$                                                      & $\mathbf{e}$                    & 5.52  \\
$\mathbf{h}_\text{pre}$                                           & $\mathbf{h}_\text{pre}$         & 69.07 \\
$\mathbf{h}_\text{post}$                                          & $\mathbf{h}_\text{post}$        & 45.26 \\
\rowcolor{ourscol} $\mathbf{h}_\text{pre}$ $\oplus$ $\mathbf{e}$  & $\mathbf{h}_\text{pre}$         & 71.65 \\
$\mathbf{h}_\text{post}$ $\oplus$ $\mathbf{e}$                    & $\mathbf{h}_\text{post}$        & 41.57 \\
\bottomrule
\end{tabular}
\end{minipage}%
\hfill
\begin{minipage}[c]{0.60\textwidth}
\captionof{table}{The impact of input and output representation for training spatial draft predictor. We train  on Janus-Pro-7B and report the score on \textbf{DPG-Bench}. We evaluate without verification ($r{=}0$) to isolate draft quality. $\mathbf{h}_\text{pre}$/$\mathbf{h}_\text{post}$: last-layer hidden state before/after the final RMSNorm. $\mathbf{e}$: token embedding. $\oplus$: concatenation. \colorbox{ourscol}{Blue} indicates our default. Predicting in discrete token space provides insufficient supervisory signal for the spatial draft prediction. Targeting pre-norm hidden states with token embedding concatenation provides the strongest learning signal.}
\label{tab:abl_design}
\end{minipage}
\end{table}

We vary the spatial draft head's input and target representations to find out the optimal feature to predict for SSD (Table~\ref{tab:abl_design}). Operating in discrete token space ($\mathbf{e} \!\to\! \mathbf{e}$) collapses to 5.52---the flat codebook distribution provides too little supervisory signal, producing outputs devoid of recognizable structure (Figure~\ref{fig:abl_design_samples}). Switching to continuous hidden-state targets recovers strong draft quality, with pre-norm states ($\mathbf{h}_\text{pre}$) consistently surpassing their post-norm counterparts ($\mathbf{h}_\text{post}$), as the unnormalized representation retains a richer feature distribution. Appending the token embedding to the input further resolves the sampling ambiguity intrinsic to greedy decoding from hidden states~\citep{li2024eagle}, achieving the strongest average score of 71.65.

\subsection{Multi-Row Anticipation Ablations}
\label{app:multi_row}

\textbf{Verification Schedule.} When anticipating multiple rows, jointly verifying all rows simultaneously causes inter-row correlations to destabilize convergence, whereas staged verification---committing the leading row before refining the trailing row---provides corrected context that yields consistently higher quality.

\begin{table}[t]
\centering
\caption{Vertical lookahead depth ablation on Lumina-mGPT-7B (\textbf{DPG-Bench}) with staged verification ($b{=}5,\; i{=}4$). $k_v$ denotes the number of rows drafted in parallel per iteration. Increasing $k_v$ reduces the number of forward passes but increases the token count per pass, so the step savings shrink quickly while the per-step cost keeps growing. $k_v{=}2$ achieves the best quality at close to the lowest latency, and is our default.}
\label{tab:abl_nvert}
\scriptsize
\begin{tabular}{@{}cIccIccIc@{}}
\toprule
\multirow{2}{*}{$k_v$} & \multirow{2}{*}{\textbf{Steps ($\downarrow$)}} & \multirow{2}{*}{\textbf{Latency ($\downarrow$)}} & \multicolumn{2}{cI}{\textbf{Acceleration ($\uparrow$)}} & \multirow{2}{*}{\textbf{DPG-Bench Average ($\uparrow$)}} \\
\cmidrule(lr){4-5}
& & & \textbf{Step} & \textbf{Latency} & \\
\midrule
AR & 2{,}352 & 96.23s & $1.00\times$ & $1.00\times$ & 76.30 \\
\midrule
1 & 303 & 14.22s & $7.76\times$ & $6.77\times$ & 73.42 \\
\rowcolor{ourscol} 2 & 280 & 13.09s & $8.40\times$ & $7.35\times$ & \textbf{74.57} \\
3 & 272 & \textbf{12.82s} & $\mathbf{8.65\times}$ & $\mathbf{7.51\times}$ & 74.31 \\
\bottomrule
\end{tabular}
\end{table}

\begin{figure}[t]
\vspace{10pt}
\centering
\includegraphics[width=\textwidth]{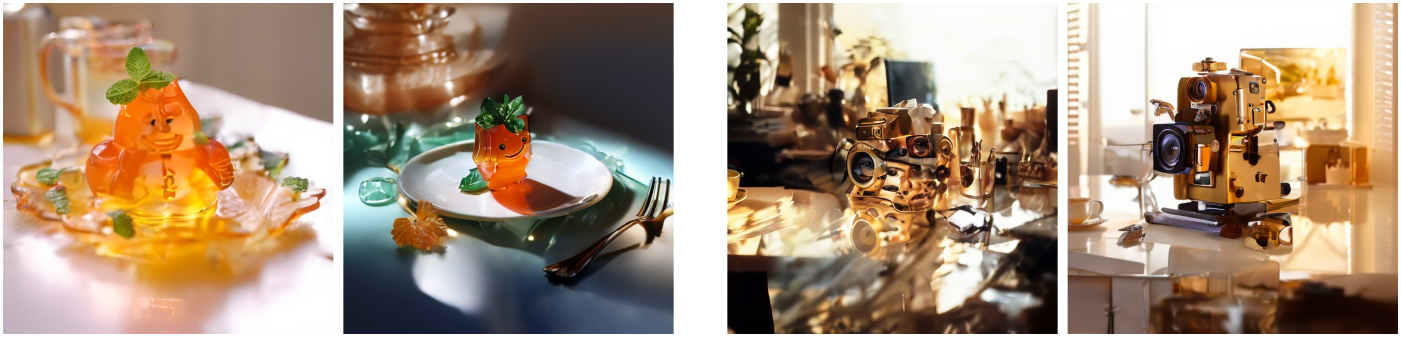}
\vspace{-3pt}
\setlength{\tabcolsep}{0pt}
\begin{tabular}{@{}p{0.24\textwidth}@{}p{0.24\textwidth}@{\hskip 12pt}p{0.24\textwidth}@{}p{0.24\textwidth}@{}}
\centering\small Joint ($b{=}9,\; i{=}0$) &
\centering\small Staged ($b{=}5,\; i{=}4$) &
\centering\small Joint ($b{=}9,\; i{=}0$) &
\centering\small Staged ($b{=}5,\; i{=}4$) \tabularnewline
\end{tabular}
\caption{Samples generated by Lumina-mGPT-7B under joint and staged verification schedules at matched total budget ($b{+}i{=}9$). Each pair compares joint verification ($b{=}9,\; i{=}0$), which processes both drafted rows simultaneously, against staged verification ($b{=}5,\; i{=}4$), which commits the first row before refining the second. Staged verification produces sharper details and more coherent spatial structure.}
\label{fig:verify_schedule_samples}
\end{figure}

To study multi-row verification, we compare joint and staged schedules at the same total verification budget ($b{+}i{=}9$, Table~\ref{tab:abl_verify_schedule}). Joint verification applies all rounds to both drafted rows simultaneously, whereas staged verification first commits the leading row and then refines the trailing row against the corrected context. Under joint verification, the leading row remains unstable across rounds and the trailing row oscillates with it, yielding a DPG-Bench score of 70.93. Staged verification reduces this coupling and improves the score to 74.57. The qualitative difference is also visible in Figure~\ref{fig:verify_schedule_samples}.

\begin{table}[t]
\centering
\caption{Verification schedule ablation on Lumina-mGPT-7B (\textbf{DPG-Bench}) with vertical lookahead $k_v{=}2$. \emph{Joint}: all $b$ verification rounds operate on both drafted rows simultaneously. \emph{Staged}: $b$ base rounds on both rows, commit the first row, then $i$ incremental rounds on the second row with the corrected first row as context. At matched total budget ($b{+}i{=}9$), staged verification outperforms joint (74.57 vs.\ 70.93) by providing corrected context before refining the trailing row.}
\label{tab:abl_verify_schedule}
\scriptsize
\begin{tabular}{@{}llIccIccIc@{}}
\toprule
\multirow{2}{*}{\textbf{Strategy}} & \multirow{2}{*}{\textbf{Schedule}} & \multirow{2}{*}{\textbf{Steps ($\downarrow$)}} & \multirow{2}{*}{\textbf{Latency ($\downarrow$)}} & \multicolumn{2}{cI}{\textbf{Acceleration ($\uparrow$)}} & \multirow{2}{*}{\textbf{DPG-Bench Average ($\uparrow$)}} \\
\cmidrule(lr){5-6}
& & & & \textbf{Step} & \textbf{Latency} & \\
\midrule
AR & -- & 2{,}352 & 96.23s & $1.00\times$ & $1.00\times$ & 76.30 \\
\midrule
Joint      & $b{=}9,\; i{=}0$ & 284 & 13.58s & $8.28\times$ & $7.09\times$ & 70.93 \\
\midrule
\multirow{4}{*}{Staged}
           & $b{=}5,\; i{=}1$ & 211 & \textbf{9.97s} & $\mathbf{11.15\times}$ & $\mathbf{9.65\times}$ & 72.22 \\
           & $b{=}5,\; i{=}2$ & 234 & 11.12s & $10.05\times$ & $8.65\times$ & 73.75 \\
           & $b{=}5,\; i{=}3$ & 257 & 12.32s & $9.15\times$ & $7.81\times$ & 73.88 \\
\rowcolor{ourscol}
           & $b{=}5,\; i{=}4$ & 280 & 13.09s & $8.40\times$ & $7.35\times$ & \textbf{74.57} \\
\bottomrule
\end{tabular}
\end{table}

\begin{figure}[t!]
\centering
\includegraphics[width=0.9\textwidth]{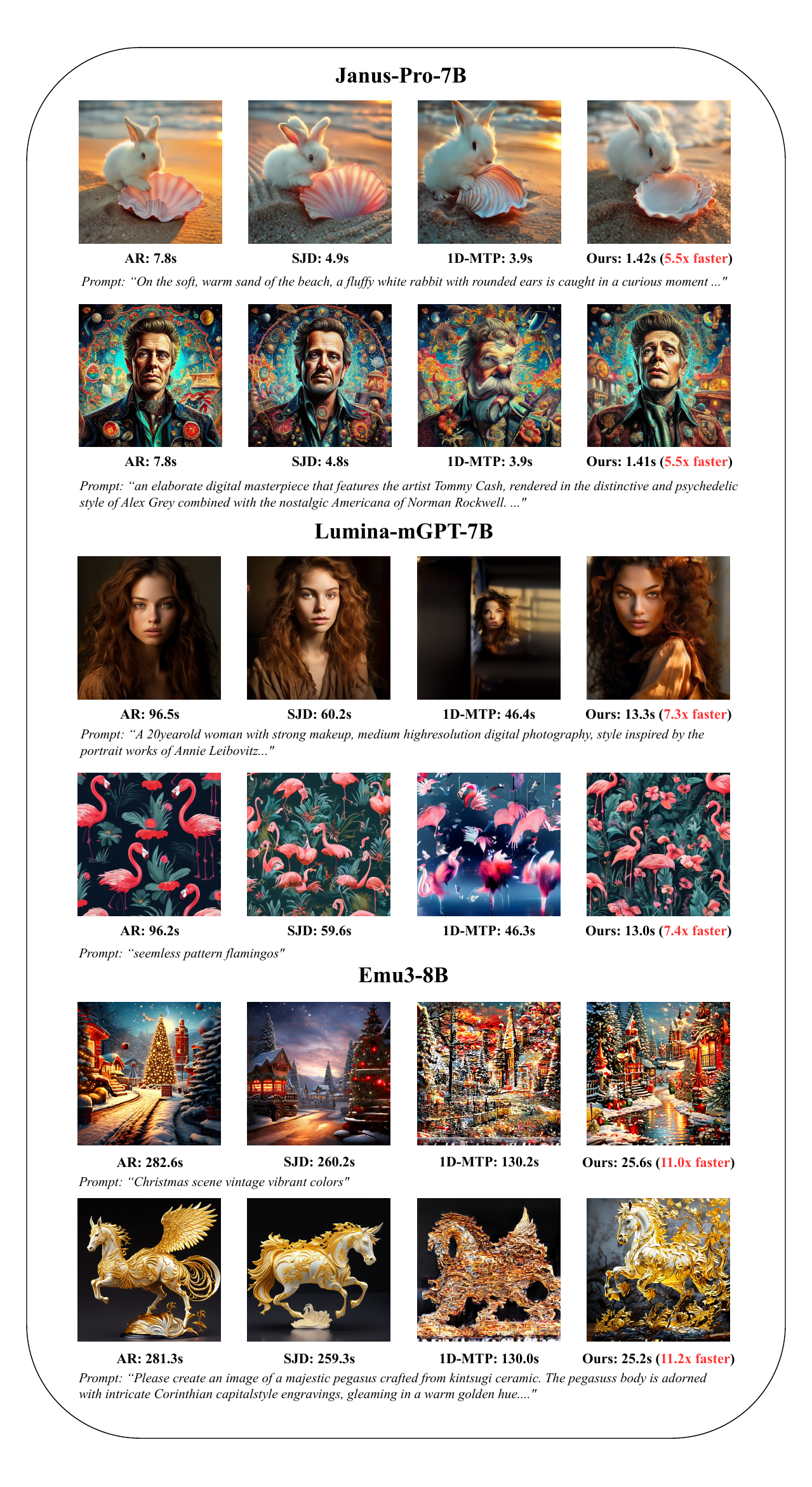}
\caption{\textbf{Extended qualitative results.} Side-by-side comparisons of AR baseline, SJD, 1D-MTP, and \model{} (Ours) across three models, demonstrating that our method achieves significant acceleration while maintaining high visual fidelity across diverse prompts.}
\label{fig:extended_vis}
\end{figure}

\subsection{Extended Visualizations}
\label{app:visualizations}

Figure~\ref{fig:extended_vis} provides extended qualitative comparisons among AR, SJD, 1D-MTP, and \model{} across all three backbones.

\subsection{Limitations}
\label{app:limitations}

Despite \model{}'s efficacy, it has several limitations. First, its relaxed parallel correction is not exactness-preserving: later positions may be verified under contexts that still contain drafted tokens, \model{} does not sample from the exact autoregressive distribution. Second, \model{} is a finite-budget approximate decoder and does not guarantee convergence to an exact fixed point; increasing the verification budget instead provides an explicit trade-off between latency and agreement with AR.

\subsection{Broader Impacts}
\label{app:broader_impacts}

\model{} substantially reduces wall-clock latency and the number of sequential forward evaluations required for autoregressive image generation, improving inference throughput. As an inference-only acceleration method, \model{} does not introduce new generative capabilities beyond those of the underlying pretrained model. Safeguard mechanisms should therefore continue to evolve alongside improvements in generation throughput.


\newpage

\end{document}